\documentclass[sigconf, nonacm]{acmart}

\AtBeginDocument{%
  }

\usepackage{booktabs} 
\usepackage{romannum}
\usepackage{graphicx}
\usepackage{booktabs}
\usepackage{amsmath}
\usepackage{url}
\usepackage{tablefootnote}
\usepackage{booktabs}
\usepackage{multirow}
\usepackage{enumitem}
\usepackage[ruled,linesnumbered]{algorithm2e}

\graphicspath{{figures/}}

\newcommand{\etal}{et al.}
\newcommand{\etc}{\textit{etc}}
\renewcommand{\paragraph}[1]{\noindent\textbf{#1.}}

\copyrightyear{2023}
\acmYear{2023}
\setcopyright{acmlicensed}
\acmConference[SA Conference Papers '23]{SIGGRAPH Asia 2023 Conference Papers}{December 12--15, 2023}{Sydney, NSW, Australia}
\acmBooktitle{SIGGRAPH Asia 2023 Conference Papers (SA Conference Papers '23), December 12--15, 2023, Sydney, NSW, Australia}
\acmPrice{15.00}
\acmDOI{10.1145/3610548.3618235}
\acmISBN{979-8-4007-0315-7/23/12}

\begin{document}

\newcommand{\x}{\mathbf{x}}
\newcommand{\y}{\mathbf{y}}
\newcommand{\s}{\mathbf{s}}
\newcommand{\w}{\mathbf{w}}
\newcommand{\cands}{\hat{\mathbf{s}}}
\newcommand{\optdist}{p_{\text{opt}}}
\newcommand{\subdist}{p_{\text{sub}}}
\newcommand{\optdata}{\mathcal{D}_{\text{opt}}}
\newcommand{\subdata}{\mathcal{D}_{\text{sub}}}
\newcommand{\vel}{\mathbf{v}}
\newcommand{\thetas}{\boldsymbol{\theta}}
\newcommand{\polys}{\mathbf{P}}
\newcommand{\loss}{\mathcal{L}}
\newcommand{\DSMloss}{\mathcal{L}_{\text{DSM}}}
\newcommand{\E}{\mathbb{E}}
\newcommand{\R}{\mathbb{R}}
\newcommand{\sspace}{\mathcal{S}}
\newcommand{\pspace}{\mathcal{P}}
\newcommand{\eps}{\epsilon}
\newcommand{\score}{\Psi_{\phi}}
\newcommand{\gscore}{\score^g}
\newcommand{\lscore}{\score^l}
\newcommand{\iscore}{\score^i}

\def\eg{\emph{e.g}.} \def\Eg{\emph{E.g}.}
\def\ie{\emph{i.e}.} \def\Ie{\emph{I.e}.}
\def\cf{\emph{c.f}.} \def\Cf{\emph{C.f}.}
\def\etc{\emph{etc}.} \def\vs{\emph{vs}.}
\def\wrt{w.r.t. } \def\dof{d.o.f. }
\def\etal{\emph{et al}. }

\title{Learning Gradient Fields for Scalable and Generalizable Irregular Packing}

\author{Tianyang Xue}
\orcid{0000-0002-1400-3566}
\authornote{Both authors contributed equally.}
\affiliation{
  \institution{Shandong University}
  \country{China}
}
\email{timhsue@gmail.com}

\author{Mingdong Wu}
\authornotemark[1]
\orcid{0009-0007-9120-4621}
\affiliation{
  \institution{Peking University}
  \country{China}
}
\email{wmingd@pku.edu.cn}

\author{Lin Lu}
\authornote{Corresponding author.}
\orcid{0000-0001-5881-892X}
\affiliation{
  \institution{Shandong University}
  \country{China}
}
\email{llu@sdu.edu.cn}

\author{Haoxuan Wang}
\orcid{0009-0003-8318-4529}
\affiliation{
  \institution{Shandong University}
  \country{China}
}

\author{Hao Dong}
\orcid{0000-0002-7984-9909}
\affiliation{
  \institution{Peking University}
  \country{China}
}
\email{hao.dong@pku.edu.cn}

\author{Baoquan Chen}
\orcid{0000-0003-4702-036X}
\affiliation{
  \institution{Peking University}
  \country{China}
}
\email{baoquan@pku.edu.cn}

\begin{abstract}

The packing problem, also known as cutting or nesting, has diverse applications in logistics, manufacturing, layout design, and atlas generation. It involves arranging irregularly shaped pieces to minimize waste while avoiding overlap. 
Recent advances in machine learning, particularly reinforcement learning, have shown promise in addressing the packing problem.
In this work, we delve deeper into a novel machine learning-based approach that formulates the packing problem as conditional generative modeling. To tackle the challenges of irregular packing, including object validity constraints and collision avoidance, our method employs the score-based diffusion model to learn a series of gradient fields. These gradient fields encode the correlations between constraint satisfaction and the spatial relationships of polygons, learned from teacher examples.
During the testing phase, packing solutions are generated using a coarse-to-fine refinement mechanism guided by the learned gradient fields. To enhance packing feasibility and optimality, we introduce two key architectural designs: multi-scale feature extraction and coarse-to-fine relation extraction.
We conduct experiments on two typical industrial packing domains, considering translations only. Empirically, our approach demonstrates spatial utilization rates comparable to, or even surpassing, those achieved by the teacher algorithm responsible for training data generation. Additionally, it exhibits some level of generalization to shape variations.
We are hopeful that this method could pave the way for new possibilities in solving the packing problem.

\end{abstract}



\begin{teaserfigure}
\includegraphics[width=\textwidth]{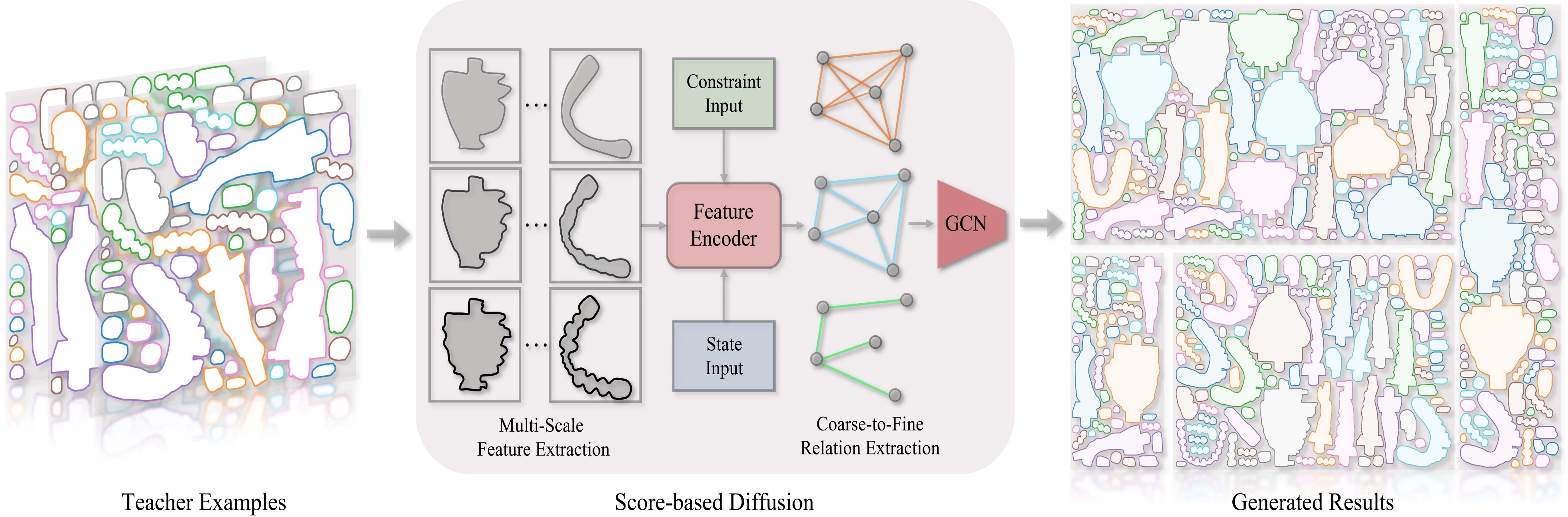}
\caption{Our goal is to model the sub-optimal packing-data distribution from teacher examples, via a score-based diffusion model designed with multi-scale feature extraction and coarse-to-fine relation extraction.
After training, we can generate solutions with unseen shapes, varying quantities, and diverse rectangular containers.}
\label{fig:teaser}    
\end{teaserfigure}

\maketitle

\section{Introduction}
\label{sec:introduction}

Irregular packing is a complex challenge in operations research and computer graphics, encompassing diverse practical applications such as logistics, transportation, manufacturing, layout design, and atlas generation~\cite{Leao2020}.
It is also known as cutting or nesting problem, which involves arranging a collection of irregularly shaped pieces onto one or multiple boards. The objective is to position each piece entirely within the board(s) without overlaps, all while minimizing the waste material produced.

One of the key applications of irregular packing is manufacturing. In these scenarios, packing problems take on different forms, such as packing parts to be cut from a larger piece of material in order to minimize waste and save costs in traditional manufacturing industries like garment manufacturing, furniture making, metal sheet cutting, or 
arranging multiple models within the build volume for simultaneous 3D printing to increase model production while minimizing material waste, and reducing print time.

In manufacturing applications, a \emph{fast}, \emph{scalable}, and \emph{generalizable} packing algorithm is crucial for real-world deployment.
\begin{itemize}
    \item Efficient algorithms streamline decision-making, simplifying deployment for end-users. 
    \item The size and aspect ratio of the container may vary depending on the raw material or machines used. 
    \item Packed objects may have similar shapes with slight deviations due to customization and personalization preferences.
\end{itemize}
These objectives motivate us to propose a learning-based method to understand the relationships and similarities among the well-packed examples from a dataset with limited size, instead of solving the packing case individually without considering any priors.

As a classic NP-hard problem~\cite{Milenkovic1999}, the irregular packing problem is typically tackled using heuristic algorithms and global optimization strategies~\cite{Guo2022}, which can be very time-consuming. 
In recent years, machine learning approaches have emerged as potential solutions to address the challenging problem of packing optimization. The success of employing reinforcement learning to address the bin packing problem~\cite{Verma2020, Zhao2020} underscores its potential for aiding in the sequential optimization of packing. While researchers have strived to enhance the initialization~\cite{Wolczyk2018} or packing sequence~\cite{Fang2023} for irregular packing, a trade-off between computational time and utilization rate persists.

To realize a fast, scalable, and generalizable packing algorithm,
our key insight is to reframe the packing problem into modeling the data distribution of sub-optimal packing examples. These examples can be generated by arbitrary packing algorithms, which we refer to as the \textit{teacher algorithms}.
Specifically, such a learning approach should learn the correlations between the packing optimality and the spatial relationships of polygons from the examples.
By modeling the local and global correlations together, the learned knowledge can generalize to unseen shapes, varying quantities, and diverse rectangular containers, which enables scalable and generalizable in-distribution packing.

During testing, a solution can be generated via several neural network inferences, without further online exploration or exhaustive optimization, which enables a fast packing process.
Thanks to the inherent parallelism of deep learning, we can generate multiple solution candidates in parallel and preserve the best one, which further improves the time efficiency of the packing process.

Inspired by \cite{Wu2022TarGF}, which proposes a score-based framework that estimates a target gradient field (TarGF) to guide object rearrangement, we aim to develop a score-based approach for modeling the teacher data. However, adapting this technique to the irregular packing problem is non-trivial due to the hard constraints on the validity of packed objects, \ie, non-overlapping. As a result, solving the irregular packing requires a coarse-to-fine refinement process guided by multiple levels of gradient fields, rather than a single gradient update (TarGF).

To achieve this goal, we introduce two key designs for the score network: \textit{multi-scale feature extraction} and \textit{coarse-to-fine relation extraction}. The former incorporates a multi-scale feature extraction network that encodes the geometric features of polygons at different scales. The latter employs a multi-layered coarse-to-fine graph convolutional network (GCN) with varying neighbor ranges for effective message-passing. By combining these two key designs, our approach tailors the score network to effectively support the coarse-to-fine refinement process, resulting in improved generalization and scalability for solving the 2D irregular packing problem.

We evaluate our method in two industrial packing domains, considering only translations. Empirical results demonstrate our method achieves spatial utilization rates that are either similar to or exceed those of teacher packing examples. Furthermore, our method scales with polygon count and generalizes to diverse container constraints and polygon shapes within the training distribution.

In summary, our contributions are listed as follows:
\begin{itemize}
[itemsep=0pt,topsep=0pt,parsep=0pt,leftmargin=15pt]
    \item We propose a score-based generative modeling approach to tackle the irregular packing problem with better efficiency, scalability, and generalization capabilities.    
    
    \item To enhance the generalization and scalability of the packing process, we introduce two designs for the score network: multi-scale feature extraction and coarse-to-fine relation extraction.
    
    \item 
    Empirical results show our method offers a trade-off between time and utilization, while also demonstrating certain scalability in terms of polygon count and in-distribution generalizability to polygon shapes.

\end{itemize}

\section{Related work}
\label{sec:related}

\subsection{Traditional packing methods}
Traditional approaches to solving the irregular packing problem involve sequential optimization and placement optimization phases~\cite{Bennell2009, Guo2022}. The sequential optimization phase determines the optimal placement order for the parts, while the placement optimization phase aims to minimize the placement gap and increase the fill rate.
Various heuristic rules, such as bottom-left (BL) and bottom-left fill (BLF), as well as mathematical models like mixed-integer linear programming, non-linear programming, and constraint programming, have been used in the placement phase~\cite{Hopper2001, Leao2020}. To place irregular shapes without overlap, researchers have proposed no-fit polygons (NFP) to describe the closest possible mutual position of two shapes~\cite{Oliveira2000, Gomes2002}.
In the sequencing phase, heuristics are combined with global search strategies such as genetic algorithms, simulated annealing, or beam search~\cite{Mavridou1997}.
Overall, traditional approaches rely on heuristics and mathematical models, which become computationally expensive with larger problems. Additionally, packing procedures must be recalculated when container or shape deviations arise.

\subsection{Packing in graphics}
The packing problem has seen extensive applications in computer graphics.
In texture atlas generation, the goal is to achieve an optimal balance between packing efficiency and minimizing distortion and boundary length in the chart mapping~\cite{Limper2018, Liu2019}. Heuristic strategies, such as one-by-one insertion with the best fit decreasing strategy or minimizing the increase on the "horizon", are employed for the chart packing process~\cite{Noell2011, Levy2002}. Another approach is to simplify the irregular packing problem by transforming the charts into rectangles~\cite{Liu2019}.

In layout design, the distribution of primitives can be optimized using techniques like generalized centroidal Voronoi tessellation~\cite{LiuCVT2009,Reinert2013, Hu2016}. 
Various heuristics, such as sample matching~\cite{Tu2022} or partial-shape matching or alignment~\cite{Kwan2016,Zou2016}, and the attract-and-repulse mechanism~\cite{Chen2017}, are employed. 
This problem setting allows primitives to scale and deform, offering more flexibility in spatial optimization, but it significantly differs from the classic irregular packing scenario.

\subsection{Packing in fabrication}
The packing problem in digital fabrications is a multidisciplinary field with applications in various manufacturing industries.
In the realm of 3D printing, researchers strive to decompose the given model into the minimum number of parts that can be fabricated within the limited build volume or to save printing costs, such as materials and time~\cite{Wang2021}. The packing processes resemble traditional methods, such as sequential optimization and heuristic placement of each part~\cite{Chen2015, Attene2015, Vanek2014}, or randomly placing all parts in a large container and then optimizing to reduce the container size~\cite{Yao2015}. The distinction lies in the collaboration between packing and model decomposition, with careful consideration of fabrication and assembly constraints during the optimization process.
In the context of laser cutting, some researchers consider material utilization in furniture design and jointly optimize part design and packing layout~\cite{Koo2017, Wu2019}.
A recent work~\cite{Cui2023} achieved state-of-the-art performance in packing generic 3D objects into the build volume by computing collision constraints in the spectral domain.

\subsection{Learning-based packing} 
Learning-based techniques for solving the packing problem have gained attention and several attempts have been devoted in recent years. 
Reinforcement learning, in particular, has been successfully employed to tackle the bin packing problem, showcasing its effectiveness in scenarios where the complete set of objects to be packed is not known in advance~\cite{Duan2019,Verma2020, Zhao2020, Zhao2021,Hu2020}. 
In terms of the irregular packing problem, \cite{Wolczyk2018} attempted to optimize the initialization, \cite{Fang2023} combined a Q-learning algorithm with a heuristic algorithm, but suffered from weak generalization and low solution efficiency.

\subsection{Score-based generative models}
Aiming to estimate the gradient of the log-likelihood of a given data distribution, the score-based generative model was first proposed by ~\cite{hyvarinen2005estimation}.
The following studies further improved the scalability of the score-based generative model by introducing a more efficient objective~\cite{denosingScoreMatching}, an annealing process~\cite{song2019generative}, or a diffusion process~\cite{song2020score}.
Inspired by the promising results, many recent works enlarged the application scope of score-based generative models, including~\cite{Wu2022TarGF, song2021solving, luo2021score, ci2022gfpose}.
Among them, \cite{Wu2022TarGF} made the first attempt to apply the score-based generative model to object rearrangement, where an agent is required to rearrange objects into a normative distribution.
Similarly, irregular packing and object rearrangement aim at obtaining a placement of objects.
However, irregular packing only requires a final packing plan and more emphasis on explicit spatial utility, while object rearrangement requires a rearrangement process in the physical world and emphasizes more on some implicit target similarity.

\section{Overview}
\label{sec:overview}

\begin{figure}[tb]
\centering
\includegraphics[width=0.9\linewidth]{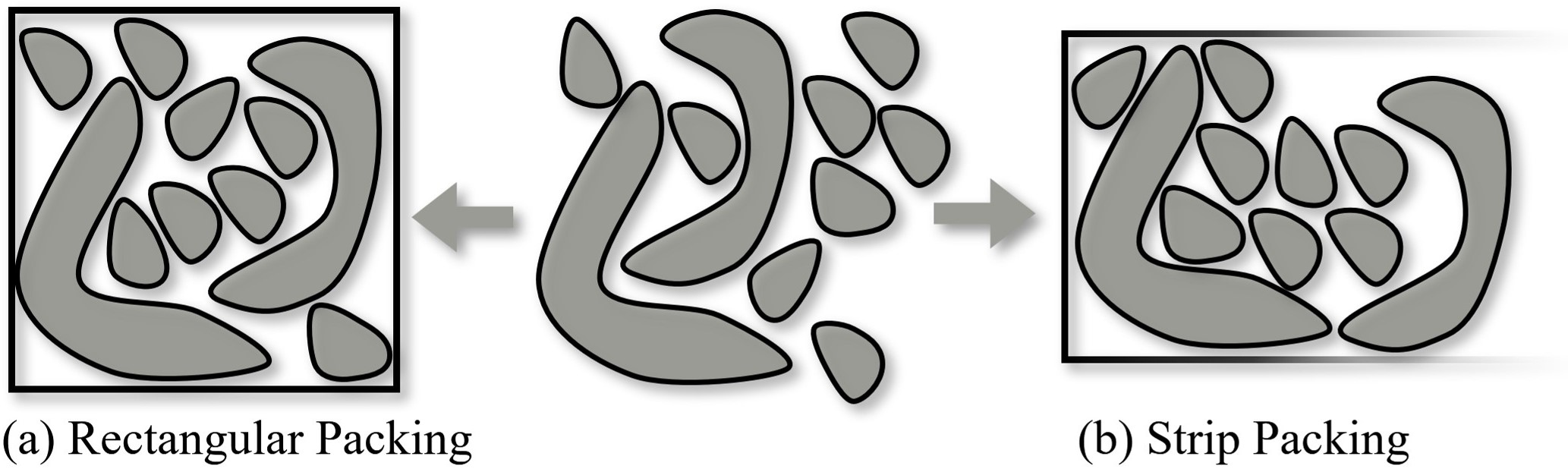}
\caption{Irregular packing problems considered in this work.}
\label{fig:problem_statment}
\end{figure} 

\begin{figure*}[tb]
\centering 
\includegraphics[width=\linewidth]{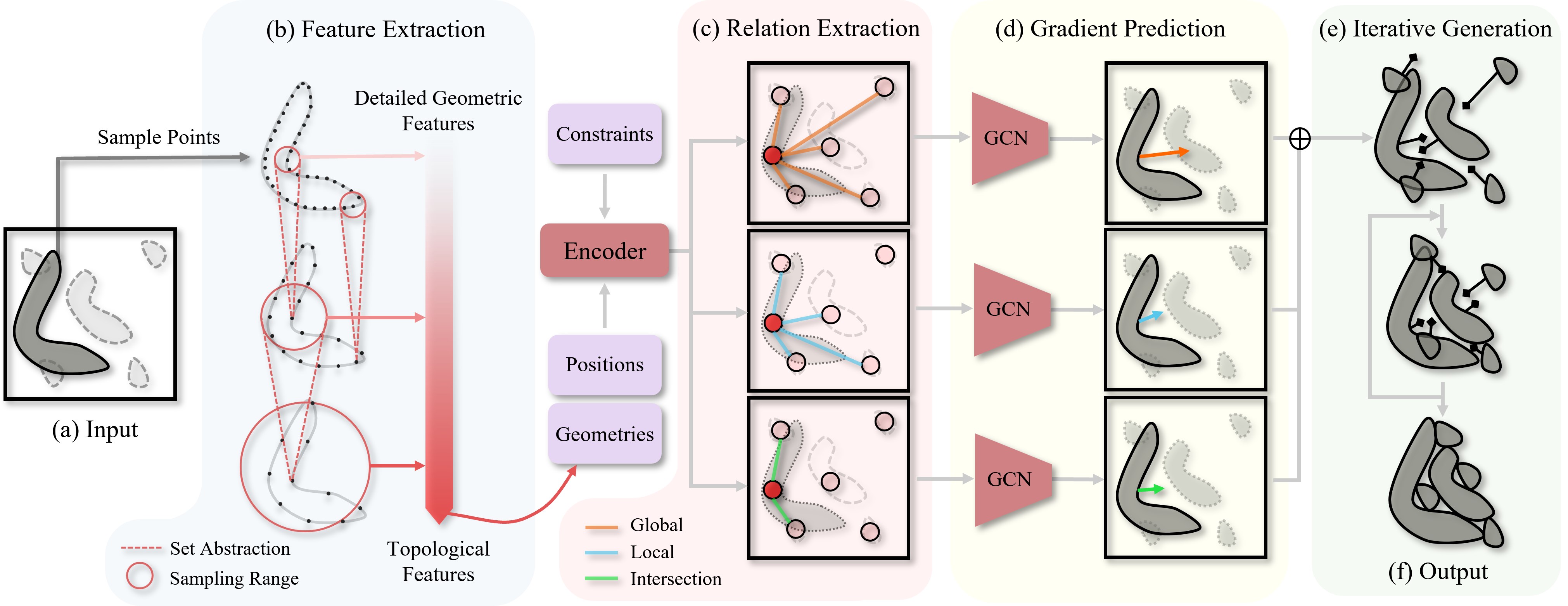}
\caption{
Our inference pipeline. 
Given an input set of polygons~(a), we extract the geometric features of polygons using a multi-scale feature extraction network~(b). 
Subsequently, we construct three relational graphs in a coarse-to-fine manner, based on the polygons' spatial information~(c). 
The relational graphs, together with the geometric, positional and constraint~(\ie, shape of container) information, are input into GCN layers to produce the aggregated gradient field (d). 
With the learned gradient fields, an output solution~(f) is generated through an iterative refinement process~(e).}
\label{fig:overview}
\end{figure*}

\subsection{Problem Statement}

The 2D irregular packing problem is a combinatorial optimization problem where the objective is to place a set of irregular polygons into a confined region without overlaps. The goal is to find the optimal placement and orientation of the polygons to maximize the usage of the packing region or minimize the overall bounding area.

In this work, we consider 2D irregular packing with only translations for simplicity. Let the polygons be denoted as $\polys = \{P_1, P_2, ..., P_n\}$, where $P_i$ is represented by the set of its points $p \in \R^2$. The packing region is denoted as $P_0$, which can be a polygonal region. For each polygon $P_i$, let $x_i, y_i$ represent the translation applied to $P_i$. The resulting translated point set is denoted as $P_i'=P_i(x_i, y_i)$.

Two polygons $P_i$ and $P_j$ are considered overlapping if the intersection of their point sets is non-empty, i.e., $P_i \cap P_j \ne \emptyset $. A polygon $P_i$ is considered out of bounds if its point set is not entirely contained within the packing region $P_0$, i.e., $P_i \not\subset P_0$. The optimal result $\s^*=(\x, \y)$ is the placement of the polygons that satisfies all the constraints while minimizing or maximizing the objective function $f(\polys, \x, \y)$ (bounding box area or space utilization). Therefore, this problem can be represented as follows:

\begin{equation}
\label{eq:define}
\begin{aligned}
    &\max\limits_{\x, \y} f(\polys, \x, \y), \\
    &\text{s.t.}
\begin{cases}
P_i' \cap P_j' = \emptyset\ &\forall i \neq j,\\
P_i' \subseteq P_0\ &\forall i.
\end{cases}
\end{aligned}
\end{equation} 

As shown in Fig. \ref{fig:problem_statment}, our task is to place multiple polygons within a finite rectangular area or an infinite strip that extends horizontally, which is a common practical scenario for most manufacturing tasks. 

\subsection{Algorithm Overview}

We propose a novel learning-based approach to tackle the problem of 2D irregular packing by formulating it as a \textit{conditional generation problem}. 
In this setting, a learning agent is tasked with modeling the conditional distribution based on a set of examples generated by a heuristic teacher algorithm~\cite{Leao2020}.

To effectively model the conditional distribution, we employ score-based diffusion models that estimate the gradient fields of the distribution by denoising diffused data examples. 
In this way, we can generate solutions for previously unseen sets of polygons in an iterative refinement process that solves a \textit{Reverse Stochastic Differential Equation}~(RSDE) or \textit{Probability Flow Ordinary Differential Equation}~(PF-ODE) guided by the learned gradient fields.

As depicted in Fig.~\ref{fig:overview}(e), our solution follows an iterative refinement process, starting from random initialization and progressively improving the solution in a coarse-to-fine manner. 
Based on this insight, we introduce two key designs for the score network: \textit{multi-scale feature extraction} and \textit{coarse-to-fine relation extraction}.

The first design~(Fig.~\ref{fig:overview}(b)) incorporates a multi-scale feature extraction network, which encodes the shapes of the polygons at different scales. This allows the network to capture fine-grained details while maintaining an understanding of the overall structure.

The second design~(Fig.~\ref{fig:overview}(c)) employs a multi-layered coarse-to-fine graph convolutional network (GCN) with varying neighbor ranges for message-passing. This GCN facilitates the extraction of relations between polygons, enabling the network to reason about their spatial dependencies in a coarse-to-fine fashion by aggregating the gradients computed by all GCN layers~(Fig.~\ref{fig:overview}(d)).

By combining these two key designs, our approach tailors the score network to effectively support the coarse-to-fine refinement process. This ensures that the network can capture both local and global information, leading to improved generalization and scalability in solving the 2D irregular packing problem.

\section{Methodology}
\label{sec:method}

\subsection{Modeling Gradients of Teacher Distribution}

We reframe 2D irregular packing as a conditional generation problem.
Ideally, the packing agent aims to estimate the implicit data distribution $\optdist$ of a set of \textit{optimal examples} $\optdata = \{(\s^*_i, \polys^*_i) \sim \optdist(\s, \polys)\}_{i=1}^{n}$ where $\s^*$ is an optimal solution of the optimization problem in Eq.~\ref{eq:define} for a given set of polygons $\polys$.
Trained on examples $\optdata$, the agent aims at estimating the conditional distribution $\optdist(\s | \polys)$.
During the test time, we can sample packing solutions from the estimated conditional distribution $\optdist(\s | \widetilde{\polys})$ for an \textit{unseen} set of polygons $\widetilde{\polys}$.
However, the exact optimal examples are hard to collect in practice. As a result, we alternatively model the implicit data distribution 
$\subdist$ of sub-optimal examples $\subdata = \{(\overline{\s^*_i}, \overline{\polys^*_i}) \sim \subdist(\s, \polys)\}_{i=1}^{n}$ where  $\overline{\s^*_i}$ is the sub-optimal solution that generated from an off-the-shelf  algorithm~\cite{Leao2020}~(\ie, the teacher) for polygons $\overline{\polys^*_i}$.

Specifically, we employ the score-based diffusion models~\cite{song2021scorebased,song2020score} to estimate the conditional data distribution $\subdist(\s | \polys)$.
The score-based diffusion models enable the learning of data distribution by introducing noise~(\ie, diffusion process) to diffuse the teacher data and subsequently learning the denoising process, which provides a powerful solution for conditional generative modeling.
Typically, we adopt Variance-Exploding~(VE) Stochastic Differential Equation~(SDE) proposed by~\cite{song2020score} to construct a continuous diffusion process $\{ \s(t) \}_{t \in [0, 1]}$ indexed by a time variable $t$ where $\s(0)\sim \subdist(\s | \polys)$ denotes the teacher solution of the polygons $\polys$. 
As the $t$ increases from 0 to 1, the time-indexed solution variable $\s(t)$ is diffused by the following SDE:
\begin{equation}
    d\s = \sqrt{\frac{d[\sigma^2(t)]}{dt}} d\mathbf{w}(t), \ 
    \sigma(t) = \sigma_{\text{min}}(\frac{\sigma_{\text{max}}}{\sigma_{\text{min}}})^t
\label{eq:forward_sde}
\end{equation}
where $\{\mathbf{w}(t)\}_{t\in [0, 1]}$ is the standard Wiener process~\cite{SDEScoreMatching}, $\sigma_{\text{min}} = 0.01$ and $\sigma_{\text{max}} = 1$.

During training, we aim to estimate the \textit{gradient fields} of the log-density, \ie, the score functions,  of the diffused conditional distributions  $\nabla_{\s} \log p_{t}(\s|\polys)$ of all $t \in [0, 1]$, where the $p_{t}(\s|\polys)$ denotes the marginal distribution of $\s(t)$:
\begin{equation}
p_{t}(\s(t)|\polys) = \int \mathcal{N}(\s(t);\s(0), \sigma^2(t)\mathbf{I}) \cdot p_0(\s(0)|\polys) \ d\s(0).
\end{equation}
Notably, when $t=0$, $p_0(\s(0)|\polys) = \subdist(\s(0)|\polys)$ is exactly the data distribution.
Thanks to the Denoising Score Matching~(DSM)~\cite{denosingScoreMatching}, we can obtain a guaranteed estimation of $\nabla_{\s} p_{t}(\s|\polys)$ by training a score network $\score: \R^{|\sspace|} \times \R^1 \times \R^{|\pspace|} \rightarrow \R^{|\sspace|}$ via the following objective:
\begin{equation}
\begin{aligned}
   \loss(\score) &= \E_{
   t\sim \mathcal{U}(\eps, 1)}
   \left[\lambda(t) \DSMloss(\score, t) \right] \\
   \DSMloss(\score, t) &= \E_{
   \s(0) \sim \subdist(\s(0)|\polys) 
   \atop 
   \s(t) \sim \mathcal{N}(\s(0), 
   \sigma^2(t)\mathbf{I})
   }
   \left[ \left\Vert\score(\s(t), t | \polys)  - \frac{\s(0) - \s(t)}{\sigma(t)^2} \right\Vert_2^2 \right]
\end{aligned}
\label{eq:score_matching_loss}
\end{equation}
where $\eps$ is a hyper-parameter that denotes the minimal noise level, $\mathcal{U}(\eps,1)$ denotes the uniform distribution on $\eps$ to 1 and $\lambda(t) = \sigma(t)^2$ is a weighting function.
When minimizing the objective in Eq.~\ref{eq:score_matching_loss}, the optimal score network satisfies $\score(\s|t, \polys) = \nabla_{\s} \log  p_{t}(\s|\polys)$ according to~\cite{denosingScoreMatching}.

\subsection{Obtaining Solutions via Iterative Refining}

\begin{figure}[tb]
\centering
\includegraphics[width=\linewidth]{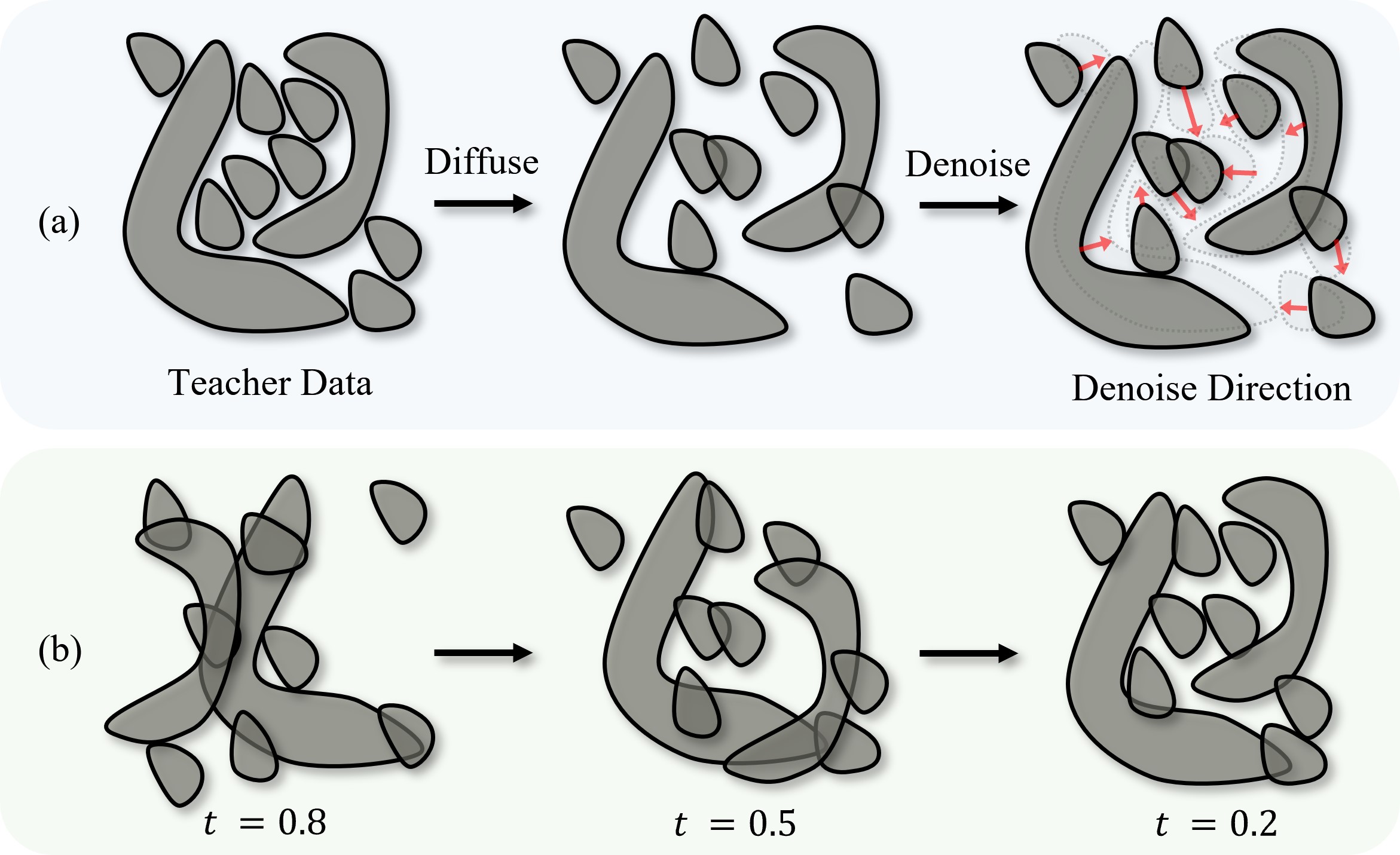}
\caption{Diffusing and denoising process. (a) 
We diffuse the teacher data with different levels of noise.
The gradient fields are trained to denoise the diffused data. 
(b) With the trained gradient fields, a solution is generated from an iterative refinement process~(\ie, the denoising process).
} 

\label{fig:noise}
\end{figure}

With the learned gradient fields, we can obtain solutions for a set of unseen polygons $\widetilde{\polys}$ by sampling from $p_{\eps}(\s|\polys)$, as $\lim_{\eps \to 0} p_{\eps}(\s|\polys) = \subdist(\s|\polys)$.
To this end, we can solve the following RSDE or PF-ODE as mentioned by~\cite{song2020score}, from $t=T_0$ to $t=\eps$.
\begin{equation}
\begin{aligned}
    d\s = - 2\sigma(t)\dot{\sigma}(t)\nabla_{\s}\log p_{t}(\s | \widetilde{\polys} ) dt + \sqrt{\frac{d[\sigma^2(t)]}{dt}} d\w(t)
\label{eq:reverse_sde}
\end{aligned}
\end{equation}
where $T_0 \in [\eps, 1]$ is a hyperparameter, $\s(T_0) \sim \mathcal{N}(\mathbf{0}, \sigma(T_0)^2\mathbf{I})$, the score functions $\nabla_{\s}\log p_{t}(\s | \polys )$ are empirically approximated by an estimated score network $\score(\s |t, \polys) \approx \nabla_{\s}\log p_{t}(\s | \polys )$.

Motivated by~\cite{Wu2022TarGF}, the output of the score network can be viewed as `pseudo velocities' imposed on the polygons $\polys$:
\begin{equation}
\label{eq:gradient}
    \score(\s) \approx \nabla_{\s}{\log{p(\s | \polys)}} = (\vel_x, \vel_y)
\end{equation}
where $(\vel_x, \vel_y)$ are linear velocities imposed on polygons' positions $\x$.
From this perspective, the stochastic process in Eq.~\ref{eq:reverse_sde} intuitively `drives' the polygons towards feasible solution via the gradient $\nabla_{\s}{\log{p(\s | \polys)}}$ and prevents them from being trapped in the local minimum via the Brownian velocities $d\w$.
By iteratively stepping the RSDE in Eq.~\ref{eq:reverse_sde}, a solution is refined from random noise to a feasible packing output in a coarse-to-fine manner.


As depicted in Fig.~\ref{fig:noise}(b), the refining process~(\ie, RSDE), necessitates the score network to assume distinct roles at different stages. 
Initially, at the onset (\eg, $t=0.8$), the polygons are positioned in appropriate configurations based on their global relationships. 
Subsequently, following the initial placement (\eg, $t=0.5$), further adjustments are made to the polygons in accordance with their local relationships. 
Finally, at the final stage (\eg, $t=0.2$), the score network is tasked with making minor adjustments to polygons that intersect with each other.

Based on this insight, the score network should be tailored to this coarse-to-fine process.
To this end, we introduce two key designs for the score network, \textit{Multi-scale Feature Extraction}~(Sec.~\ref{subsec:feature_extraction}) and \textit{Coarse-to-fine Relation Extraction}~(Sec.~\ref{subsec:relation_extraction}).

\label{subsec:train_and_gen}

\subsection{Multi-scale Feature Extraction}
\label{subsec:feature_extraction}

The packing problem requires the incorporation of extensive geometric information, which includes the topological contours and local geometric nuances of the polygons.

Taking inspiration from identifying objects of different sizes in object detection problems, our model employs a multi-scale perception network.
Initially, we employed Feature Pyramid Networks (FPN)~\cite{lin2017fpn} for capturing multi-scale information. 
However, employing the FPN necessitates the mapping of each polygon onto a two-dimensional image, leading to significant parameter overhead and degradation of fine geometric details.

We incorporated PointNets~\cite{qi2017pointnet, qi2017pointnetpp, Qian2022PointNeXt} architecture (Fig. \ref{fig:overview}(b)). 
With a U-shape architecture, PointNets enable the processing of point set information at various scales through \emph{set abstracting} procedures and different sampling resolutions.
Empirically, we observed that PointNeXt demonstrates remarkable proficiency in extracting geometric features from polygons~(Sec.~\ref{subsec:validation}). 

Apart from the shape information of the polygons, position and constraint information are also required.
The constraints and positions are encoded via an MLP encoder, together with the geometric information, into feature vectors $f_c$, $f_p$, and $f_g$. These feature vectors are utilized by the subsequent layer of the network (Fig. \ref{fig:overview}(b-c)).

\subsection{Coarse-to-fine Relation Extraction}
\label{subsec:relation_extraction}

When packing polygons, determining which polygons should be grouped together, or separated, based on their geometric properties poses a highly challenging task. 
Specifically, it involves the precise analysis of the spatial distribution of polygons and the complex relationships between their shapes and positions, which requires effective message-passing between the polygons.
GCNs~\cite{zhao2022PCT} are capable of effectively handling objects with complex relationships and learning the interactions between them, making them highly suitable for addressing polygon packing problems. 

\begin{figure}[b]
\centering
\includegraphics[width=\linewidth]{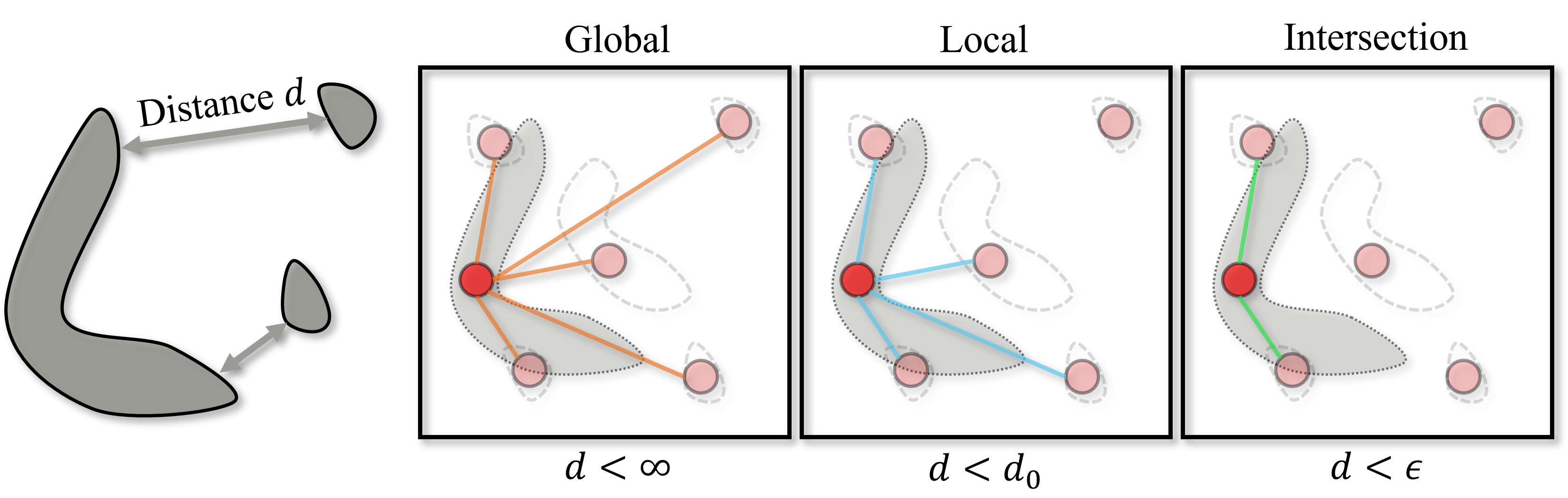}
\caption{
We build relational graphs on polygons in a coarse-to-fine manner, where connectivity is based on varying distance thresholds. The Gilbert–Johnson–Keerthi algorithm \cite{Montanari2017gjk, Montanari2018gjk} is applied to compute the distances between polygons.
}
\label{fig:distance}
\end{figure} 

Relationships between polygons are encoded as edges into GCNs (Fig. \ref{fig:overview}(c)). 
During generation, the optimal placement under the current polygon set necessitates a fully connected graph for robust set perception, forming the \emph{global layer} $\Psi_{\phi}^g$. 
However, varying polygon numbers alter global layer perception, influencing model generalization. 
Thus, we designed a \emph{local layer} $\Psi_{\phi}^l$ to perceive polygons within a small scope, enhancing model generalizability. 
Furthermore, an \emph{intersection layer} $\Psi_{\phi}^i$ was designed to prevent polygon overlap. 
In particular, the edges of polygons in the graph network are determined by their distances (Fig. \ref{fig:distance}).
We assign a unique identifier $i$ to each polygon, with its corresponding features denoted as $f^i=(f_p^i,f_g^i,f_c)$. 
The set of edges $E$ is defined such that if there is a connection between two polygons, $i$ and $j$, then $(i, j) \in E$.

We encode these connectivity relationships using dynamic GCN \cite{wang2019dynamic}. 
Moreover, to augment the network's discernment of interrelationships among polygons, positional information is manifested as relative form.
In the aggregation layer of the graph network, the aggregation of feature is expressed as

\begin{equation}
    {f^i}' = \max_{j : (i,j) \in E} h_\Theta(f^i, f^j) = \max_{j : (i,j) \in E} h_\theta(f^i | f^i - f^j)
\end{equation}
while the aggregation of positional information is expressed as
\begin{equation}
    h_\Theta^0(f^i, f^j) = h_\theta(f^i - f^j).
\end{equation}

\begin{figure}[tb]
\centering
\includegraphics[width=1.0\linewidth]{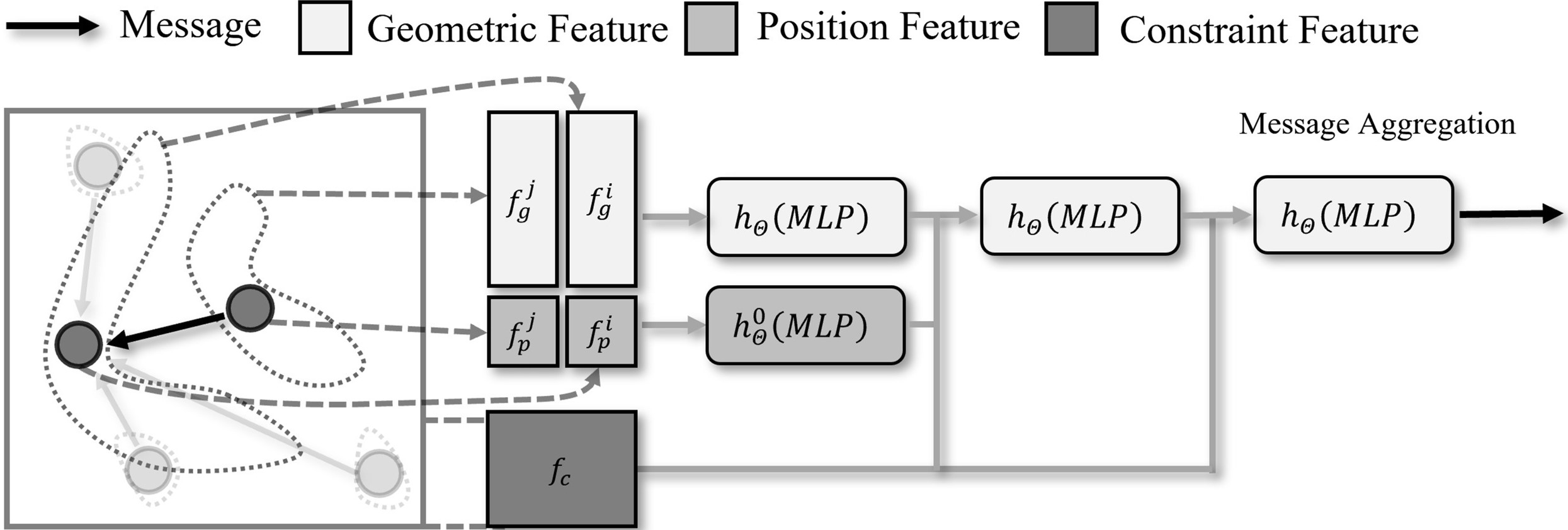}
\caption{Implementations of Message-passing layer. }
\label{fig:network}
\end{figure}

All the features are aggregated to the connected nodes through the connectivity scheme in Fig. \ref{fig:distance} and the convolutional kernels illustrated in Fig. \ref{fig:network}. The GCNs subsequently produce the gradient fields for each graph (Fig. \ref{fig:overview}(d)). Then the final output gradient field is obtained from mean-pooling.

\section{Results and Discussion}
\label{sec:results}

\subsection{Dataset}
\label{subsec:dataset}
\begin{figure}[tb]
\centering
\includegraphics[width=1.0\linewidth]{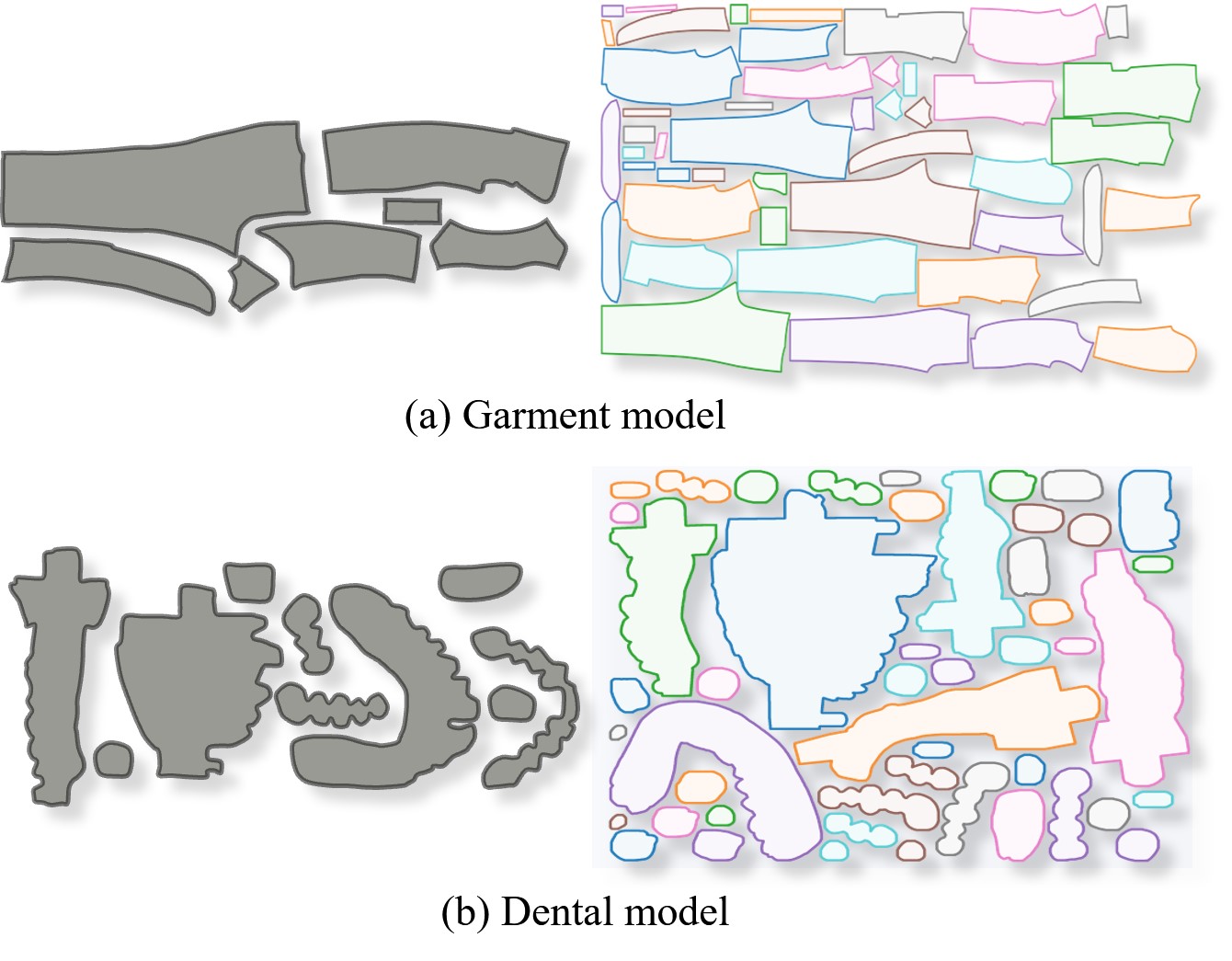}
\caption{
The left side shows polygon examples, while the right side displays the teacher dataset. The garment dataset consists of 314 polygons, ranging from simple to complex shapes. The most complex polygon comprises 284 points, whereas the simplest one has 5 points. In contrast, the dental model dataset includes 109 polygons with varying levels of complexity, featuring irregular shapes and vertex counts ranging from 52 to 1735. 
}
\label{fig:teacher_data}
\end{figure}

In this paper, we conduct experiments on two datasets obtained from real manufacturing environments~(Fig. \ref{fig:teacher_data}). The first dataset is an additive manufacturing dataset that encompasses various dental models. The second dataset focuses on fabric cutting and manufacturing, primarily composed of garment pieces.

To generate the teacher datasets, we implemented a bottom-left-fill algorithm based on a preprocessed No-Fit Polygon and further optimized it using simulated annealing~\cite{Leao2020}. 

From each dataset, 70\% of the data were randomly selected and used as the training set, while the remaining 30\% was designated as the test set to assess the model's in-distribution generalization ability of shapes. 
Subsequently, 48 polygons were randomly chosen from each training set, and approximately 20,000 sets of polygons were generated. The strip height or container dimensions during packing were selected from a range of 1280 to 1920.

\subsection{Implementation}
\label{subsec:implementation}
The polygons are sampled into 256 points, which serve as the input to a modified version of \emph{PointNeXt}.
As the layers progress deeper, the number of sampled points decreases by one-fourth in each subsequent layer. The receptive field of each layer encompasses the 16 or 8 nearest neighboring points surrounding each sampled point. The network's final output is passed through a head layer, which produces a geometric feature vector $f_g$ of length 128.

The relation extraction network consists of coarse-to-fine GCNs mentioned in Sec.~\ref{subsec:relation_extraction}. 
A fully connected graph is used to construct the \emph{global layer}.
The \emph{local layer} emphasizes local information, where we establish connections among polygons that have a distance $d$ smaller than $d_0=200$.
Lastly, in the \emph{intersection layer}, we only connect those polygons that intersect.  
During the training process, $t \in [0,1]$. 
The loss function is designed as follows:
\begin{equation}
    \loss_{\text{all}}(\phi) = \loss(\gscore) + \loss(\lscore) + \loss(\iscore) + \loss(\score)
\end{equation}  
where $\score$ expresses the coupling of the three gradient fields by summing them and $\loss$ is similar to Eq.~\ref{eq:score_matching_loss}. 
This design plays a crucial role in maintaining a balance of losses across distinct layers and their interconnected outcomes throughout the training process.

To stabilize the training, we normalize all the 2D coordinates~(\eg, the point clouds $\polys$ and the solution $[\x, \y]$) by multiplying $\frac{1}{2000}$.

The PF-ODE is solved by the Euler method. The RSDE is solved by the Euler-Maruyama method. We defer the detailed pseudo codes of iterative refinement in supplementary materials.

\subsection{Validation}
\label{subsec:validation}


We validate our method in terms of efficiency, scalability, and generalization as follows.

\subsubsection{Efficiency} 
We created a test dataset containing sets of 48 polygons, with strip heights or container dimensions ranging from 1280 to 1920. 
As shown in Tab. \ref{tab:efficiency_result}, our model could generate results faster than the teacher algorithms. Moreover, it achieves comparable utilization rates to teacher algorithms and even surpasses their utilization in all garment data.


\subsubsection{Scalability for the number of polygons} 
To assess the scalability of our model with respect to the number of polygons, we created a test dataset with polygon counts ranging from 20 to 128. We randomly set the strip heights between 1280 and 1920.
As shown in Fig. \ref{fig:scalability}, our algorithm displays scalability over different polygon counts, maintaining a steady spatial utilization rate as the polygon count grows. While our algorithm doesn't enforce strict non-collision constraints, it's worth noting that some resulting solutions might be unfeasible. This is a key factor influencing the scalability of our model concerning varying numbers of polygons.

\subsubsection{Generalization for container constraints} 
We conducted experiments to assess the generalization capability of our model with respect to container constraints. 
These experiments utilized a test dataset comprising 48 polygons and strip heights ranging from 640 to 3920. 
The obtained results, illustrated in Fig. \ref{fig:scalability}, clearly indicate the model's ability to generalize and conform to the container boundaries, within the training distribution.

\subsubsection{Generalization for polygon shapes}
Our model's generalization capability towards polygon shapes was demonstrated through the validation performed on 30\% of unseen test data.
To further validate the learned geometric features, we conducted experiments where we altered the local geometric structures and the overall topological structures of the test data, as illustrated in Fig.~\ref{fig:shape}.
The results of these experiments showcased our model's enhanced generalization, affirming the effectiveness of its learned geometric features.

\subsection{Evaluation}
\label{subsec:evaluation}

\begin{table}
\caption{Comparison of Time Consumption and Space Utilization between our method with different parameters and alternative packing methods. Herein, $b$ denotes batch size, while $E_s$ represents the number of Euler method steps. In our generation, we set $T_0=0.8$ (Garment dataset) and $T_0=0.6$ (Dental dataset).
We performed our algorithm, teacher packer and rectangular packer on the same Linux server equipped with an AMD EPYC 7T83 CPU and an NVIDIA RTX3090 GPU.
Additionally, xatlas was executed on a Windows server outfitted with an I7-13700K CPU.}
\begin{tabular}{ccccc}
\hline
Dataset            & \multicolumn{2}{c}{Garment} & \multicolumn{2}{c}{Dental} \\ \cline{2-5} 
Algo.               & Util (\%)     & Time (s)    & Util (\%)    & Time (s)    \\ \hline
Ours ($b$=128, $E_s$=64)  & 65.72         & 6.03        & 60.13        & 6.22       \\
Ours ($b$=512, $E_s$=64)  & 66.94         & 21.91       & 62.10        & 23.12       \\
Ours ($b$=512, $E_s$=128) & 67.66         & 45.22       & 62.44        & 46.32       \\
Ours ($b$=1024, $E_s$=128)& 68.16         & 86.34       & 63.20        & 88.43       \\
xatlas Packer\tablefootnote{\url{https://github.com/jpcy/xatlas}}            
                          & 64.75         & 1.24        & 64.80        & 1.82        \\
Teacher Packer            & 64.22         & 42.32       & 65.65        & 48.65       \\
Rectangular Packer\tablefootnote{\url{https://github.com/nothings/stb/blob/master/stb_rect_pack.h}}        
                          & 62.04         & 0.0005      & 50.64        & 0.0004      \\ \hline
\end{tabular}
\label{tab:efficiency_result}
\end{table}

In pursuit of a comprehensive understanding of the model, we conducted a qualitative analysis of the generated gradient field. Concretely, we utilized a pair of identical polygons extracted from a teacher test dataset for the construction of the gradient field. The visual representations, as depicted in Figure \ref{fig:gradient_field}, unequivocally demonstrate the proficient capacity of the trained gradient field to navigate us toward the optimal position. Noteworthy is the consistent guidance provided by the gradient to polygons located beyond the boundaries, steering them toward the interior region. Furthermore, our observations indicate that manipulating the value of $t$ to smaller magnitudes induces the polygons to generate gradients directed toward unoccupied spaces.

To ascertain the effectiveness of our coarse-to-fine network design, we conducted ablation studies using a test dataset comprising 48 polygons. We maintained the activation of the \emph{global layer} and systematically evaluated the network's performance with either the \emph{local layer} or the \emph{intersection layer} individually enabled. Through these experimental investigations (see Fig.~\ref{fig:ablation}), we elucidated the pivotal role played by the \emph{local layer} in producing generalized outcomes. Furthermore, the \emph{intersection layer} demonstrated its capacity to effectively acquire knowledge of constraints, leading to the successful elimination of polygon overlap.

\section{Conclusion}

In this study, we introduce a novel approach that employs score-based generative modeling to address 2D irregular packing problems, focusing solely on translations. Unlike reinforcement-learning-based methods, we derive gradient fields from the teacher data distribution to produce packing solutions. Utilizing these trained gradient fields, we generate solutions through a coarse-to-fine iterative refinement process. To improve packing feasibility and optimality, we incorporate two essential network designs: multi-scale feature extraction and coarse-to-fine relation extraction. Empirical findings highlight our approach's impressive space utilization. Furthermore, our method exhibits scalability concerning polygon count and generalizes effectively to various container constraints and polygon shapes within the training distribution.

We summarise the main limitations and future work as follows.

In this work, while our primary focus is on irregular packing problems, we have not extensively explored three-dimensional datasets. 3D packing problems present unique challenges due to physical constraints, making it difficult to obtain gradient fields. However, we believe our method naturally extends to three-dimensional data for a few reasons. Firstly, our feature extraction network~(\ie, PointNets) was designed for processing three-dimensional point clouds. 
Furthermore, we can specify additional conditions for the network to account for constraints like packing stability~\cite{Wang2019}, and support structures minimization~\cite{Cao2021}, beyond just the container constraint.

Our method employs neural networks, which face the challenge of ensuring that all outputs are overlap-free (Fig. \ref{fig:scalability}), mainly due to the difficulty in incorporating strict non-overlapping constraints. Integrating traditional packing methods may help address this challenge.

Our method is constrained by sub-optimal training data stemming from conditional generative modeling. Nevertheless, we observe that our generated outputs surpass the quality of the training data, indicating the possibility of augmenting our approach with reinforcement learning to explore optimal solution spaces. Additionally, incorporating rotational aspects into our framework could further enhance optimality. Thus, we may consider integrating an orientational perception module into our method for this purpose.

We also confirmed the feasibility of capturing boundary features using Denoising Score Matching. This opens up the possibility of nesting irregular containers by encoding irregular boundaries into the condition.

Geometric features are crucial in polygon extraction, particularly in pair-wise operations. While PointNets are feasible for packing problems, enhancing the network's generalization capabilities can be achieved by augmenting the feature extraction network with traditional methods that assist in extracting geometric descriptors.

\begin{acks}
We thank all the anonymous reviewers for their valuable comments and constructive suggestions. 
Thanks for the dental dataset provided by Shining 3D and garment dataset from Aliyun. 
This work is supported in part by grants from NSFC (61972232, 62376006), the National Youth Talent Support Program (8200800081), and Shenzhen Collaborative Innovation Program 
(CJGJZD20200617102202007,\\ CJGJZD2021048092601003).
\end{acks}

\bibliographystyle{ACM-Reference-Format}
\bibliography{reference}


\begin{thebibliography}{54}


\ifx \showCODEN    \undefined \def \showCODEN     #1{\unskip}     \fi
\ifx \showDOI      \undefined \def \showDOI       #1{#1}\fi
\ifx \showISBNx    \undefined \def \showISBNx     #1{\unskip}     \fi
\ifx \showISBNxiii \undefined \def \showISBNxiii  #1{\unskip}     \fi
\ifx \showISSN     \undefined \def \showISSN      #1{\unskip}     \fi
\ifx \showLCCN     \undefined \def \showLCCN      #1{\unskip}     \fi
\ifx \shownote     \undefined \def \shownote      #1{#1}          \fi
\ifx \showarticletitle \undefined \def \showarticletitle #1{#1}   \fi
\ifx \showURL      \undefined \def \showURL       {\relax}        \fi
\providecommand\bibfield[2]{#2}
\providecommand\bibinfo[2]{#2}
\providecommand\natexlab[1]{#1}
\providecommand\showeprint[2][]{arXiv:#2}

\bibitem[Attene(2015)]%
        {Attene2015}
\bibfield{author}{\bibinfo{person}{Marco Attene}.} \bibinfo{year}{2015}\natexlab{}.
\newblock \showarticletitle{Shapes In a Box: Disassembling 3D Objects for Efficient Packing and Fabrication}.
\newblock \bibinfo{journal}{\emph{Computer Graphics Forum}} \bibinfo{volume}{34}, \bibinfo{number}{8} (\bibinfo{date}{may} \bibinfo{year}{2015}), \bibinfo{pages}{64--76}.
\newblock
\urldef\tempurl%
\url{https://doi.org/10.1111/cgf.12608}
\showDOI{\tempurl}


\bibitem[Bennell and Oliveira(2009)]%
        {Bennell2009}
\bibfield{author}{\bibinfo{person}{J~A Bennell} {and} \bibinfo{person}{J~F Oliveira}.} \bibinfo{year}{2009}\natexlab{}.
\newblock \showarticletitle{A tutorial in irregular shape packing problems}.
\newblock \bibinfo{journal}{\emph{Journal of the Operational Research Society}} \bibinfo{volume}{60}, \bibinfo{number}{sup1} (\bibinfo{date}{may} \bibinfo{year}{2009}), \bibinfo{pages}{S93--S105}.
\newblock
\urldef\tempurl%
\url{https://doi.org/10.1057/jors.2008.169}
\showDOI{\tempurl}


\bibitem[Cao et~al\mbox{.}(2021)]%
        {Cao2021}
\bibfield{author}{\bibinfo{person}{Lingxin Cao}, \bibinfo{person}{Lihao Tian}, \bibinfo{person}{Hao Peng}, \bibinfo{person}{Yu Zhou}, {and} \bibinfo{person}{Lin Lu}.} \bibinfo{year}{2021}\natexlab{}.
\newblock \showarticletitle{Constrained stacking in DLP 3D printing}.
\newblock \bibinfo{journal}{\emph{Computers \& Graphics}}  \bibinfo{volume}{95} (\bibinfo{year}{2021}), \bibinfo{pages}{60--68}.
\newblock
\showISSN{0097-8493}
\urldef\tempurl%
\url{https://doi.org/10.1016/j.cag.2021.01.003}
\showDOI{\tempurl}


\bibitem[Chen et~al\mbox{.}(2017)]%
        {Chen2017}
\bibfield{author}{\bibinfo{person}{Weikai Chen}, \bibinfo{person}{Yuexin Ma}, \bibinfo{person}{Sylvain Lefebvre}, \bibinfo{person}{Shiqing Xin}, \bibinfo{person}{Jon{\`{a}}s Mart{\'{\i}}nez}, {and} \bibinfo{person}{Wenping Wang}.} \bibinfo{year}{2017}\natexlab{}.
\newblock \showarticletitle{Fabricable tile decors}.
\newblock \bibinfo{journal}{\emph{{ACM} Transactions on Graphics}} \bibinfo{volume}{36}, \bibinfo{number}{6} (\bibinfo{date}{nov} \bibinfo{year}{2017}), \bibinfo{pages}{1--15}.
\newblock
\urldef\tempurl%
\url{https://doi.org/10.1145/3130800.3130817}
\showDOI{\tempurl}


\bibitem[Chen et~al\mbox{.}(2015)]%
        {Chen2015}
\bibfield{author}{\bibinfo{person}{Xuelin Chen}, \bibinfo{person}{Hao Zhang}, \bibinfo{person}{Jinjie Lin}, \bibinfo{person}{Ruizhen Hu}, \bibinfo{person}{Lin Lu}, \bibinfo{person}{Qixing Huang}, \bibinfo{person}{Bedrich Benes}, \bibinfo{person}{Daniel Cohen-Or}, {and} \bibinfo{person}{Baoquan Chen}.} \bibinfo{year}{2015}\natexlab{}.
\newblock \showarticletitle{Dapper: decompose-and-pack for 3D printing}.
\newblock \bibinfo{journal}{\emph{{ACM} Transactions on Graphics}} \bibinfo{volume}{34}, \bibinfo{number}{6} (\bibinfo{date}{nov} \bibinfo{year}{2015}), \bibinfo{pages}{1--12}.
\newblock
\urldef\tempurl%
\url{https://doi.org/10.1145/2816795.2818087}
\showDOI{\tempurl}


\bibitem[Ci et~al\mbox{.}(2022)]%
        {ci2022gfpose}
\bibfield{author}{\bibinfo{person}{Hai Ci}, \bibinfo{person}{Mingdong Wu}, \bibinfo{person}{Wentao Zhu}, \bibinfo{person}{Xiaoxuan Ma}, \bibinfo{person}{Hao Dong}, \bibinfo{person}{Fangwei Zhong}, {and} \bibinfo{person}{Yizhou Wang}.} \bibinfo{year}{2022}\natexlab{}.
\newblock \bibinfo{title}{GFPose: Learning 3D Human Pose Prior with Gradient Fields}.
\newblock
\newblock
\showeprint[arxiv]{2212.08641}~[cs.CV]


\bibitem[Cui et~al\mbox{.}(2023)]%
        {Cui2023}
\bibfield{author}{\bibinfo{person}{Qiaodong Cui}, \bibinfo{person}{Victor Rong}, \bibinfo{person}{Desai Chen}, {and} \bibinfo{person}{Wojciech Matusik}.} \bibinfo{year}{2023}\natexlab{}.
\newblock \showarticletitle{Dense, Interlocking-Free and Scalable Spectral Packing of Generic 3D Objects}.
\newblock \bibinfo{journal}{\emph{ACM Trans. Graph.}} \bibinfo{volume}{42}, \bibinfo{number}{4}, Article \bibinfo{articleno}{141} (\bibinfo{date}{jul} \bibinfo{year}{2023}), \bibinfo{numpages}{14}~pages.
\newblock
\showISSN{0730-0301}
\urldef\tempurl%
\url{https://doi.org/10.1145/3592126}
\showDOI{\tempurl}


\bibitem[Duan et~al\mbox{.}(2019)]%
        {Duan2019}
\bibfield{author}{\bibinfo{person}{Lu Duan}, \bibinfo{person}{Haoyuan Hu}, \bibinfo{person}{Yu Qian}, \bibinfo{person}{Yu Gong}, \bibinfo{person}{Xiaodong Zhang}, \bibinfo{person}{Jiangwen Wei}, {and} \bibinfo{person}{Yinghui Xu}.} \bibinfo{year}{2019}\natexlab{}.
\newblock \showarticletitle{A Multi-Task Selected Learning Approach for Solving 3D Flexible Bin Packing Problem}. In \bibinfo{booktitle}{\emph{Proceedings of the 18th International Conference on Autonomous Agents and MultiAgent Systems}} (Montreal QC, Canada) \emph{(\bibinfo{series}{AAMAS '19})}. \bibinfo{publisher}{International Foundation for Autonomous Agents and Multiagent Systems}, \bibinfo{address}{Richland, SC}, \bibinfo{pages}{1386–1394}.
\newblock
\showISBNx{9781450363099}


\bibitem[Fang et~al\mbox{.}(2023)]%
        {Fang2023}
\bibfield{author}{\bibinfo{person}{Jie Fang}, \bibinfo{person}{Yunqing Rao}, \bibinfo{person}{Xusheng Zhao}, {and} \bibinfo{person}{Bing Du}.} \bibinfo{year}{2023}\natexlab{}.
\newblock \showarticletitle{A Hybrid Reinforcement Learning Algorithm for 2D Irregular Packing Problems}.
\newblock \bibinfo{journal}{\emph{Mathematics}} \bibinfo{volume}{11}, \bibinfo{number}{2} (\bibinfo{date}{jan} \bibinfo{year}{2023}), \bibinfo{pages}{327}.
\newblock
\urldef\tempurl%
\url{https://doi.org/10.3390/math11020327}
\showDOI{\tempurl}


\bibitem[Gomes and Oliveira(2002)]%
        {Gomes2002}
\bibfield{author}{\bibinfo{person}{A.Miguel Gomes} {and} \bibinfo{person}{Jos{\'{e}}~F. Oliveira}.} \bibinfo{year}{2002}\natexlab{}.
\newblock \showarticletitle{A 2-exchange heuristic for nesting problems}.
\newblock \bibinfo{journal}{\emph{European Journal of Operational Research}} \bibinfo{volume}{141}, \bibinfo{number}{2} (\bibinfo{date}{sep} \bibinfo{year}{2002}), \bibinfo{pages}{359--370}.
\newblock
\urldef\tempurl%
\url{https://doi.org/10.1016/s0377-2217(02)00130-3}
\showDOI{\tempurl}


\bibitem[Guo et~al\mbox{.}(2022)]%
        {Guo2022}
\bibfield{author}{\bibinfo{person}{Baosu Guo}, \bibinfo{person}{Yu Zhang}, \bibinfo{person}{Jingwen Hu}, \bibinfo{person}{Jinrui Li}, \bibinfo{person}{Fenghe Wu}, \bibinfo{person}{Qingjin Peng}, {and} \bibinfo{person}{Quan Zhang}.} \bibinfo{year}{2022}\natexlab{}.
\newblock \showarticletitle{Two-dimensional irregular packing problems: A review}.
\newblock \bibinfo{journal}{\emph{Frontiers in Mechanical Engineering}}  \bibinfo{volume}{8} (\bibinfo{date}{aug} \bibinfo{year}{2022}).
\newblock
\urldef\tempurl%
\url{https://doi.org/10.3389/fmech.2022.966691}
\showDOI{\tempurl}


\bibitem[Hopper and Turton(2001)]%
        {Hopper2001}
\bibfield{author}{\bibinfo{person}{E Hopper} {and} \bibinfo{person}{B.C.H Turton}.} \bibinfo{year}{2001}\natexlab{}.
\newblock \showarticletitle{An empirical investigation of meta-heuristic and heuristic algorithms for a 2D packing problem}.
\newblock \bibinfo{journal}{\emph{European Journal of Operational Research}} \bibinfo{volume}{128}, \bibinfo{number}{1} (\bibinfo{date}{jan} \bibinfo{year}{2001}), \bibinfo{pages}{34--57}.
\newblock
\urldef\tempurl%
\url{https://doi.org/10.1016/s0377-2217(99)00357-4}
\showDOI{\tempurl}


\bibitem[Hu et~al\mbox{.}(2020)]%
        {Hu2020}
\bibfield{author}{\bibinfo{person}{Ruizhen Hu}, \bibinfo{person}{Juzhan Xu}, \bibinfo{person}{Bin Chen}, \bibinfo{person}{Minglun Gong}, \bibinfo{person}{Hao Zhang}, {and} \bibinfo{person}{Hui Huang}.} \bibinfo{year}{2020}\natexlab{}.
\newblock \showarticletitle{{TAP}-Net: transport-and-pack using reinforcement learning}.
\newblock \bibinfo{journal}{\emph{{ACM} Transactions on Graphics}} \bibinfo{volume}{39}, \bibinfo{number}{6} (\bibinfo{date}{nov} \bibinfo{year}{2020}), \bibinfo{pages}{1--15}.
\newblock
\urldef\tempurl%
\url{https://doi.org/10.1145/3414685.3417796}
\showDOI{\tempurl}


\bibitem[Hu et~al\mbox{.}(2016)]%
        {Hu2016}
\bibfield{author}{\bibinfo{person}{Wenchao Hu}, \bibinfo{person}{Zhonggui Chen}, \bibinfo{person}{Hao Pan}, \bibinfo{person}{Yizhou Yu}, \bibinfo{person}{Eitan Grinspun}, {and} \bibinfo{person}{Wenping Wang}.} \bibinfo{year}{2016}\natexlab{}.
\newblock \showarticletitle{Surface Mosaic Synthesis with Irregular Tiles}.
\newblock \bibinfo{journal}{\emph{{IEEE} Transactions on Visualization and Computer Graphics}} \bibinfo{volume}{22}, \bibinfo{number}{3} (\bibinfo{date}{mar} \bibinfo{year}{2016}), \bibinfo{pages}{1302--1313}.
\newblock
\urldef\tempurl%
\url{https://doi.org/10.1109/tvcg.2015.2498620}
\showDOI{\tempurl}


\bibitem[Hyv\"{a}rinen(2005)]%
        {hyvarinen2005estimation}
\bibfield{author}{\bibinfo{person}{Aapo Hyv\"{a}rinen}.} \bibinfo{year}{2005}\natexlab{}.
\newblock \showarticletitle{Estimation of Non-Normalized Statistical Models by Score Matching}.
\newblock \bibinfo{journal}{\emph{J. Mach. Learn. Res.}}  \bibinfo{volume}{6} (\bibinfo{date}{dec} \bibinfo{year}{2005}), \bibinfo{pages}{695–709}.
\newblock
\showISSN{1532-4435}


\bibitem[Koo et~al\mbox{.}(2017)]%
        {Koo2017}
\bibfield{author}{\bibinfo{person}{Bongjin Koo}, \bibinfo{person}{Jean Hergel}, \bibinfo{person}{Sylvain Lefebvre}, {and} \bibinfo{person}{Niloy~J. Mitra}.} \bibinfo{year}{2017}\natexlab{}.
\newblock \showarticletitle{Towards Zero-Waste Furniture Design}.
\newblock \bibinfo{journal}{\emph{{IEEE} Transactions on Visualization and Computer Graphics}} \bibinfo{volume}{23}, \bibinfo{number}{12} (\bibinfo{date}{dec} \bibinfo{year}{2017}), \bibinfo{pages}{2627--2640}.
\newblock
\urldef\tempurl%
\url{https://doi.org/10.1109/tvcg.2016.2633519}
\showDOI{\tempurl}


\bibitem[Kwan et~al\mbox{.}(2016)]%
        {Kwan2016}
\bibfield{author}{\bibinfo{person}{Kin~Chung Kwan}, \bibinfo{person}{Lok~Tsun Sinn}, \bibinfo{person}{Chu Han}, \bibinfo{person}{Tien-Tsin Wong}, {and} \bibinfo{person}{Chi-Wing Fu}.} \bibinfo{year}{2016}\natexlab{}.
\newblock \showarticletitle{Pyramid of arclength descriptor for generating collage of shapes}.
\newblock \bibinfo{journal}{\emph{{ACM} Transactions on Graphics}} \bibinfo{volume}{35}, \bibinfo{number}{6} (\bibinfo{date}{nov} \bibinfo{year}{2016}), \bibinfo{pages}{1--12}.
\newblock
\urldef\tempurl%
\url{https://doi.org/10.1145/2980179.2980234}
\showDOI{\tempurl}


\bibitem[Leao et~al\mbox{.}(2020)]%
        {Leao2020}
\bibfield{author}{\bibinfo{person}{Aline~A.S. Leao}, \bibinfo{person}{Franklina~M.B. Toledo}, \bibinfo{person}{Jos{\'{e}}~Fernando Oliveira}, \bibinfo{person}{Maria~Ant{\'{o}}nia Carravilla}, {and} \bibinfo{person}{Ram{\'{o}}n Alvarez-Vald{\'{e}}s}.} \bibinfo{year}{2020}\natexlab{}.
\newblock \showarticletitle{Irregular packing problems: A review of mathematical models}.
\newblock \bibinfo{journal}{\emph{European Journal of Operational Research}} \bibinfo{volume}{282}, \bibinfo{number}{3} (\bibinfo{date}{may} \bibinfo{year}{2020}), \bibinfo{pages}{803--822}.
\newblock
\urldef\tempurl%
\url{https://doi.org/10.1016/j.ejor.2019.04.045}
\showDOI{\tempurl}


\bibitem[L{\'{e}}vy et~al\mbox{.}(2002)]%
        {Levy2002}
\bibfield{author}{\bibinfo{person}{Bruno L{\'{e}}vy}, \bibinfo{person}{Sylvain Petitjean}, \bibinfo{person}{Nicolas Ray}, {and} \bibinfo{person}{J{\'{e}}rome Maillot}.} \bibinfo{year}{2002}\natexlab{}.
\newblock \showarticletitle{Least squares conformal maps for automatic texture atlas generation}.
\newblock \bibinfo{journal}{\emph{{ACM} Transactions on Graphics}} \bibinfo{volume}{21}, \bibinfo{number}{3} (\bibinfo{date}{jul} \bibinfo{year}{2002}), \bibinfo{pages}{362--371}.
\newblock
\urldef\tempurl%
\url{https://doi.org/10.1145/566654.566590}
\showDOI{\tempurl}


\bibitem[Limper et~al\mbox{.}(2018)]%
        {Limper2018}
\bibfield{author}{\bibinfo{person}{Max Limper}, \bibinfo{person}{Nicholas Vining}, {and} \bibinfo{person}{ALLA SHEFFER}.} \bibinfo{year}{2018}\natexlab{}.
\newblock \showarticletitle{Box cutter: atlas refinement for efficient packing via void elimination}.
\newblock \bibinfo{journal}{\emph{{ACM} Transactions on Graphics}} \bibinfo{volume}{37}, \bibinfo{number}{4} (\bibinfo{date}{jul} \bibinfo{year}{2018}), \bibinfo{pages}{1--13}.
\newblock
\urldef\tempurl%
\url{https://doi.org/10.1145/3197517.3201328}
\showDOI{\tempurl}


\bibitem[Lin et~al\mbox{.}(2017)]%
        {lin2017fpn}
\bibfield{author}{\bibinfo{person}{Tsung-Yi Lin}, \bibinfo{person}{Piotr Dollár}, \bibinfo{person}{Ross Girshick}, \bibinfo{person}{Kaiming He}, \bibinfo{person}{Bharath Hariharan}, {and} \bibinfo{person}{Serge Belongie}.} \bibinfo{year}{2017}\natexlab{}.
\newblock \bibinfo{title}{Feature Pyramid Networks for Object Detection}.
\newblock
\newblock
\showeprint[arxiv]{1612.03144}~[cs.CV]


\bibitem[Liu et~al\mbox{.}(2019)]%
        {Liu2019}
\bibfield{author}{\bibinfo{person}{Hao-Yu Liu}, \bibinfo{person}{Xiao-Ming Fu}, \bibinfo{person}{Chunyang Ye}, \bibinfo{person}{Shuangming Chai}, {and} \bibinfo{person}{Ligang Liu}.} \bibinfo{year}{2019}\natexlab{}.
\newblock \showarticletitle{Atlas refinement with bounded packing efficiency}.
\newblock \bibinfo{journal}{\emph{{ACM} Transactions on Graphics}} \bibinfo{volume}{38}, \bibinfo{number}{4} (\bibinfo{date}{jul} \bibinfo{year}{2019}), \bibinfo{pages}{1--13}.
\newblock
\urldef\tempurl%
\url{https://doi.org/10.1145/3306346.3323001}
\showDOI{\tempurl}


\bibitem[Liu et~al\mbox{.}(2009)]%
        {LiuCVT2009}
\bibfield{author}{\bibinfo{person}{Yang Liu}, \bibinfo{person}{Wenping Wang}, \bibinfo{person}{Bruno L\'{e}vy}, \bibinfo{person}{Feng Sun}, \bibinfo{person}{Dong-Ming Yan}, \bibinfo{person}{Lin Lu}, {and} \bibinfo{person}{Chenglei Yang}.} \bibinfo{year}{2009}\natexlab{}.
\newblock \showarticletitle{On Centroidal Voronoi Tessellation—energy Smoothness and Fast Computation}.
\newblock \bibinfo{journal}{\emph{ACM Trans. Graph.}} \bibinfo{volume}{28}, \bibinfo{number}{4}, Article \bibinfo{articleno}{101} (\bibinfo{date}{sep} \bibinfo{year}{2009}), \bibinfo{numpages}{17}~pages.
\newblock
\showISSN{0730-0301}
\urldef\tempurl%
\url{https://doi.org/10.1145/1559755.1559758}
\showDOI{\tempurl}


\bibitem[Luo and Hu(2021)]%
        {luo2021score}
\bibfield{author}{\bibinfo{person}{Shitong Luo} {and} \bibinfo{person}{Wei Hu}.} \bibinfo{year}{2021}\natexlab{}.
\newblock \showarticletitle{Score-based point cloud denoising}. In \bibinfo{booktitle}{\emph{Proceedings of the IEEE/CVF International Conference on Computer Vision}}. \bibinfo{publisher}{IEEE Computer Society}, \bibinfo{address}{Los Alamitos, CA, USA}, \bibinfo{pages}{4583--4592}.
\newblock


\bibitem[Mavridou and Pardalos(1997)]%
        {Mavridou1997}
\bibfield{author}{\bibinfo{person}{Thelma~D. Mavridou} {and} \bibinfo{person}{Panos~M. Pardalos}.} \bibinfo{year}{1997}\natexlab{}.
\newblock \showarticletitle{Simulated Annealing and Genetic Algorithms for the Facility Layout Problem: A Survey}.
\newblock \bibinfo{journal}{\emph{Computational Optimization and Applications}} \bibinfo{volume}{7}, \bibinfo{number}{1} (\bibinfo{year}{1997}), \bibinfo{pages}{111--126}.
\newblock
\urldef\tempurl%
\url{https://doi.org/10.1023/a:1008623913524}
\showDOI{\tempurl}


\bibitem[Milenkovic(1999)]%
        {Milenkovic1999}
\bibfield{author}{\bibinfo{person}{Victor~J. Milenkovic}.} \bibinfo{year}{1999}\natexlab{}.
\newblock \showarticletitle{Rotational polygon containment and minimum enclosure using only robust 2D constructions}.
\newblock \bibinfo{journal}{\emph{Computational Geometry}} \bibinfo{volume}{13}, \bibinfo{number}{1} (\bibinfo{date}{may} \bibinfo{year}{1999}), \bibinfo{pages}{3--19}.
\newblock
\urldef\tempurl%
\url{https://doi.org/10.1016/s0925-7721(99)00006-1}
\showDOI{\tempurl}


\bibitem[Montanari and Petrinic(2018)]%
        {Montanari2018gjk}
\bibfield{author}{\bibinfo{person}{Mattia Montanari} {and} \bibinfo{person}{Nik Petrinic}.} \bibinfo{year}{2018}\natexlab{}.
\newblock \showarticletitle{{OpenGJK} for C, C{\#}~and Matlab: Reliable solutions to distance queries between convex bodies in three-dimensional space}.
\newblock \bibinfo{journal}{\emph{{SoftwareX}}}  \bibinfo{volume}{7} (\bibinfo{date}{jan} \bibinfo{year}{2018}), \bibinfo{pages}{352--355}.
\newblock
\urldef\tempurl%
\url{https://doi.org/10.1016/j.softx.2018.10.002}
\showDOI{\tempurl}


\bibitem[Montanari et~al\mbox{.}(2017)]%
        {Montanari2017gjk}
\bibfield{author}{\bibinfo{person}{Mattia Montanari}, \bibinfo{person}{Nik Petrinic}, {and} \bibinfo{person}{Ettore Barbieri}.} \bibinfo{year}{2017}\natexlab{}.
\newblock \showarticletitle{Improving the {GJK} Algorithm for Faster and More Reliable Distance Queries Between Convex Objects}.
\newblock \bibinfo{journal}{\emph{{ACM} Transactions on Graphics}} \bibinfo{volume}{36}, \bibinfo{number}{4} (\bibinfo{date}{jun} \bibinfo{year}{2017}), \bibinfo{pages}{1}.
\newblock
\urldef\tempurl%
\url{https://doi.org/10.1145/3072959.3083724}
\showDOI{\tempurl}


\bibitem[Nöll and Strieker(2011)]%
        {Noell2011}
\bibfield{author}{\bibinfo{person}{T. Nöll} {and} \bibinfo{person}{D. Strieker}.} \bibinfo{year}{2011}\natexlab{}.
\newblock \showarticletitle{Efficient Packing of Arbitrary Shaped Charts for Automatic Texture Atlas Generation}.
\newblock \bibinfo{journal}{\emph{Computer Graphics Forum}} \bibinfo{volume}{30}, \bibinfo{number}{4} (\bibinfo{date}{jun} \bibinfo{year}{2011}), \bibinfo{pages}{1309--1317}.
\newblock
\urldef\tempurl%
\url{https://doi.org/10.1111/j.1467-8659.2011.01990.x}
\showDOI{\tempurl}


\bibitem[Oliveira et~al\mbox{.}(2000)]%
        {Oliveira2000}
\bibfield{author}{\bibinfo{person}{Jos{\'{e}}~F. Oliveira}, \bibinfo{person}{A.~Miguel Gomes}, {and} \bibinfo{person}{J.~Soeiro Ferreira}.} \bibinfo{year}{2000}\natexlab{}.
\newblock \showarticletitle{{TOPOS} {\textendash} A new constructive algorithm for nesting problems}.
\newblock \bibinfo{journal}{\emph{{OR} Spektrum}} \bibinfo{volume}{22}, \bibinfo{number}{2} (\bibinfo{year}{2000}), \bibinfo{pages}{263}.
\newblock
\urldef\tempurl%
\url{https://doi.org/10.1007/s002910050105}
\showDOI{\tempurl}


\bibitem[Qi et~al\mbox{.}(2017a)]%
        {qi2017pointnet}
\bibfield{author}{\bibinfo{person}{Charles~R. Qi}, \bibinfo{person}{Hao Su}, \bibinfo{person}{Kaichun Mo}, {and} \bibinfo{person}{Leonidas~J. Guibas}.} \bibinfo{year}{2017}\natexlab{a}.
\newblock \bibinfo{title}{PointNet: Deep Learning on Point Sets for 3D Classification and Segmentation}.
\newblock
\newblock
\showeprint[arxiv]{1612.00593}~[cs.CV]


\bibitem[Qi et~al\mbox{.}(2017b)]%
        {qi2017pointnetpp}
\bibfield{author}{\bibinfo{person}{Charles~R. Qi}, \bibinfo{person}{Li Yi}, \bibinfo{person}{Hao Su}, {and} \bibinfo{person}{Leonidas~J. Guibas}.} \bibinfo{year}{2017}\natexlab{b}.
\newblock \bibinfo{title}{PointNet++: Deep Hierarchical Feature Learning on Point Sets in a Metric Space}.
\newblock
\newblock
\showeprint[arxiv]{1706.02413}~[cs.CV]


\bibitem[Qian et~al\mbox{.}(2022)]%
        {Qian2022PointNeXt}
\bibfield{author}{\bibinfo{person}{Guocheng Qian}, \bibinfo{person}{Yuchen Li}, \bibinfo{person}{Houwen Peng}, \bibinfo{person}{Jinjie Mai}, \bibinfo{person}{Hasan Hammoud}, \bibinfo{person}{Mohamed Elhoseiny}, {and} \bibinfo{person}{Bernard Ghanem}.} \bibinfo{year}{2022}\natexlab{}.
\newblock \showarticletitle{PointNeXt: Revisiting PointNet++ with Improved Training and Scaling Strategies}. In \bibinfo{booktitle}{\emph{Advances in Neural Information Processing Systems (NeurIPS)}}.
\newblock


\bibitem[Reinert et~al\mbox{.}(2013)]%
        {Reinert2013}
\bibfield{author}{\bibinfo{person}{Bernhard Reinert}, \bibinfo{person}{Tobias Ritschel}, {and} \bibinfo{person}{Hans-Peter Seidel}.} \bibinfo{year}{2013}\natexlab{}.
\newblock \showarticletitle{Interactive by-example design of artistic packing layouts}.
\newblock \bibinfo{journal}{\emph{{ACM} Transactions on Graphics}} \bibinfo{volume}{32}, \bibinfo{number}{6} (\bibinfo{date}{nov} \bibinfo{year}{2013}), \bibinfo{pages}{1--7}.
\newblock
\urldef\tempurl%
\url{https://doi.org/10.1145/2508363.2508409}
\showDOI{\tempurl}


\bibitem[Song and Ermon(2019)]%
        {song2019generative}
\bibfield{author}{\bibinfo{person}{Yang Song} {and} \bibinfo{person}{Stefano Ermon}.} \bibinfo{year}{2019}\natexlab{}.
\newblock \showarticletitle{Generative modeling by estimating gradients of the data distribution}.
\newblock \bibinfo{journal}{\emph{Advances in Neural Information Processing Systems}}  \bibinfo{volume}{32} (\bibinfo{year}{2019}).
\newblock


\bibitem[Song et~al\mbox{.}(2021a)]%
        {song2021solving}
\bibfield{author}{\bibinfo{person}{Yang Song}, \bibinfo{person}{Liyue Shen}, \bibinfo{person}{Lei Xing}, {and} \bibinfo{person}{Stefano Ermon}.} \bibinfo{year}{2021}\natexlab{a}.
\newblock \showarticletitle{Solving inverse problems in medical imaging with score-based generative models}.
\newblock \bibinfo{journal}{\emph{arXiv preprint arXiv:2111.08005}} (\bibinfo{year}{2021}).
\newblock


\bibitem[Song et~al\mbox{.}(2020)]%
        {SDEScoreMatching}
\bibfield{author}{\bibinfo{person}{Yang Song}, \bibinfo{person}{Jascha Sohl-Dickstein}, \bibinfo{person}{Diederik~P Kingma}, \bibinfo{person}{Abhishek Kumar}, \bibinfo{person}{Stefano Ermon}, {and} \bibinfo{person}{Ben Poole}.} \bibinfo{year}{2020}\natexlab{}.
\newblock \showarticletitle{Score-based generative modeling through stochastic differential equations}.
\newblock \bibinfo{journal}{\emph{arXiv preprint arXiv:2011.13456}} (\bibinfo{year}{2020}).
\newblock


\bibitem[Song et~al\mbox{.}(2021b)]%
        {song2020score}
\bibfield{author}{\bibinfo{person}{Yang Song}, \bibinfo{person}{Jascha Sohl-Dickstein}, \bibinfo{person}{Diederik~P. Kingma}, \bibinfo{person}{Abhishek Kumar}, \bibinfo{person}{Stefano Ermon}, {and} \bibinfo{person}{Ben Poole}.} \bibinfo{year}{2021}\natexlab{b}.
\newblock \bibinfo{title}{Score-Based Generative Modeling through Stochastic Differential Equations}.
\newblock
\newblock
\showeprint[arxiv]{2011.13456}~[cs.LG]


\bibitem[Song et~al\mbox{.}(2021c)]%
        {song2021scorebased}
\bibfield{author}{\bibinfo{person}{Yang Song}, \bibinfo{person}{Jascha Sohl-Dickstein}, \bibinfo{person}{Diederik~P. Kingma}, \bibinfo{person}{Abhishek Kumar}, \bibinfo{person}{Stefano Ermon}, {and} \bibinfo{person}{Ben Poole}.} \bibinfo{year}{2021}\natexlab{c}.
\newblock \bibinfo{title}{Score-Based Generative Modeling through Stochastic Differential Equations}.
\newblock
\newblock
\showeprint[arxiv]{2011.13456}~[cs.LG]


\bibitem[Tu et~al\mbox{.}(2022)]%
        {Tu2022}
\bibfield{author}{\bibinfo{person}{Peihan Tu}, \bibinfo{person}{Li-Yi Wei}, {and} \bibinfo{person}{Matthias Zwicker}.} \bibinfo{year}{2022}\natexlab{}.
\newblock \showarticletitle{Clustered vector textures}.
\newblock \bibinfo{journal}{\emph{{ACM} Transactions on Graphics}} \bibinfo{volume}{41}, \bibinfo{number}{4} (\bibinfo{date}{jul} \bibinfo{year}{2022}), \bibinfo{pages}{1--23}.
\newblock
\urldef\tempurl%
\url{https://doi.org/10.1145/3528223.3530062}
\showDOI{\tempurl}


\bibitem[Vanek et~al\mbox{.}(2014)]%
        {Vanek2014}
\bibfield{author}{\bibinfo{person}{J. Vanek}, \bibinfo{person}{J.~A.~Garcia Galicia}, \bibinfo{person}{B. Benes}, \bibinfo{person}{R. M{\v{e}}ch}, \bibinfo{person}{N. Carr}, \bibinfo{person}{O. Stava}, {and} \bibinfo{person}{G.~S. Miller}.} \bibinfo{year}{2014}\natexlab{}.
\newblock \showarticletitle{{PackMerger}: A 3D Print Volume Optimizer}.
\newblock \bibinfo{journal}{\emph{Computer Graphics Forum}} \bibinfo{volume}{33}, \bibinfo{number}{6} (\bibinfo{date}{may} \bibinfo{year}{2014}), \bibinfo{pages}{322--332}.
\newblock
\urldef\tempurl%
\url{https://doi.org/10.1111/cgf.12353}
\showDOI{\tempurl}


\bibitem[Verma et~al\mbox{.}(2020)]%
        {Verma2020}
\bibfield{author}{\bibinfo{person}{Richa Verma}, \bibinfo{person}{Aniruddha Singhal}, \bibinfo{person}{Harshad Khadilkar}, \bibinfo{person}{Ansuma Basumatary}, \bibinfo{person}{Siddharth Nayak}, \bibinfo{person}{Harsh~Vardhan Singh}, \bibinfo{person}{Swagat Kumar}, {and} \bibinfo{person}{Rajesh Sinha}.} \bibinfo{year}{2020}\natexlab{}.
\newblock \bibinfo{title}{A Generalized Reinforcement Learning Algorithm for Online 3D Bin-Packing}.
\newblock
\newblock
\showeprint[arxiv]{2007.00463}~[cs.AI]


\bibitem[Vincent(2011)]%
        {denosingScoreMatching}
\bibfield{author}{\bibinfo{person}{Pascal Vincent}.} \bibinfo{year}{2011}\natexlab{}.
\newblock \showarticletitle{A connection between score matching and denoising autoencoders}.
\newblock \bibinfo{journal}{\emph{Neural computation}} \bibinfo{volume}{23}, \bibinfo{number}{7} (\bibinfo{year}{2011}), \bibinfo{pages}{1661--1674}.
\newblock


\bibitem[Wang and Hauser(2019)]%
        {Wang2019}
\bibfield{author}{\bibinfo{person}{Fan Wang} {and} \bibinfo{person}{Kris Hauser}.} \bibinfo{year}{2019}\natexlab{}.
\newblock \showarticletitle{Stable Bin Packing of Non-Convex 3D Objects with a Robot Manipulator}. In \bibinfo{booktitle}{\emph{2019 International Conference on Robotics and Automation (ICRA)}} (Montreal, QC, Canada). \bibinfo{publisher}{IEEE Press}, \bibinfo{pages}{8698–8704}.
\newblock
\urldef\tempurl%
\url{https://doi.org/10.1109/ICRA.2019.8794049}
\showDOI{\tempurl}


\bibitem[Wang et~al\mbox{.}(2019)]%
        {wang2019dynamic}
\bibfield{author}{\bibinfo{person}{Yue Wang}, \bibinfo{person}{Yongbin Sun}, \bibinfo{person}{Ziwei Liu}, \bibinfo{person}{Sanjay~E. Sarma}, \bibinfo{person}{Michael~M. Bronstein}, {and} \bibinfo{person}{Justin~M. Solomon}.} \bibinfo{year}{2019}\natexlab{}.
\newblock \bibinfo{title}{Dynamic Graph CNN for Learning on Point Clouds}.
\newblock
\newblock
\showeprint[arxiv]{1801.07829}~[cs.CV]


\bibitem[Wang et~al\mbox{.}(2021)]%
        {Wang2021}
\bibfield{author}{\bibinfo{person}{Ziqi Wang}, \bibinfo{person}{Peng Song}, {and} \bibinfo{person}{Mark Pauly}.} \bibinfo{year}{2021}\natexlab{}.
\newblock \showarticletitle{State of the Art on Computational Design of Assemblies with Rigid Parts}.
\newblock \bibinfo{journal}{\emph{Computer Graphics Forum}} \bibinfo{volume}{40}, \bibinfo{number}{2} (\bibinfo{date}{may} \bibinfo{year}{2021}), \bibinfo{pages}{633--657}.
\newblock
\urldef\tempurl%
\url{https://doi.org/10.1111/cgf.142660}
\showDOI{\tempurl}


\bibitem[Wo{\l}czyk(2018)]%
        {Wolczyk2018}
\bibfield{author}{\bibinfo{person}{Maciej Wo{\l}czyk}.} \bibinfo{year}{2018}\natexlab{}.
\newblock \showarticletitle{Deep learning-based initialization for object packing}.
\newblock \bibinfo{journal}{\emph{Schedae Informaticae}}  \bibinfo{volume}{27} (\bibinfo{year}{2018}), \bibinfo{pages}{9--17}.
\newblock
\urldef\tempurl%
\url{https://doi.org/10.4467/20838476si.18.001.10406}
\showDOI{\tempurl}


\bibitem[Wu et~al\mbox{.}(2019)]%
        {Wu2019}
\bibfield{author}{\bibinfo{person}{Chenming Wu}, \bibinfo{person}{Haisen Zhao}, \bibinfo{person}{Chandrakana Nandi}, \bibinfo{person}{Jeffrey~I. Lipton}, \bibinfo{person}{Zachary Tatlock}, {and} \bibinfo{person}{Adriana Schulz}.} \bibinfo{year}{2019}\natexlab{}.
\newblock \showarticletitle{Carpentry compiler}.
\newblock \bibinfo{journal}{\emph{{ACM} Transactions on Graphics}} \bibinfo{volume}{38}, \bibinfo{number}{6} (\bibinfo{date}{nov} \bibinfo{year}{2019}), \bibinfo{pages}{1--14}.
\newblock
\urldef\tempurl%
\url{https://doi.org/10.1145/3355089.3356518}
\showDOI{\tempurl}


\bibitem[Wu et~al\mbox{.}(2022)]%
        {Wu2022TarGF}
\bibfield{author}{\bibinfo{person}{Mingdong Wu}, \bibinfo{person}{Fangwei Zhong}, \bibinfo{person}{Yulong Xia}, {and} \bibinfo{person}{Hao Dong}.} \bibinfo{year}{2022}\natexlab{}.
\newblock \showarticletitle{TarGF: Learning Target Gradient Field to Rearrange Objects without Explicit Goal Specification}. In \bibinfo{booktitle}{\emph{Advances in Neural Information Processing Systems}}, \bibfield{editor}{\bibinfo{person}{S.~Koyejo}, \bibinfo{person}{S.~Mohamed}, \bibinfo{person}{A.~Agarwal}, \bibinfo{person}{D.~Belgrave}, \bibinfo{person}{K.~Cho}, {and} \bibinfo{person}{A.~Oh}} (Eds.), Vol.~\bibinfo{volume}{35}. \bibinfo{publisher}{Curran Associates, Inc.}, \bibinfo{pages}{31986--31999}.
\newblock
\urldef\tempurl%
\url{https://proceedings.neurips.cc/paper_files/paper/2022/file/cf5a019ae9c11b4be88213ce3f85d85c-Paper-Conference.pdf}
\showURL{%
\tempurl}


\bibitem[Yao et~al\mbox{.}(2015)]%
        {Yao2015}
\bibfield{author}{\bibinfo{person}{Miaojun Yao}, \bibinfo{person}{Zhili Chen}, \bibinfo{person}{Linjie Luo}, \bibinfo{person}{Rui Wang}, {and} \bibinfo{person}{Huamin Wang}.} \bibinfo{year}{2015}\natexlab{}.
\newblock \showarticletitle{Level-set-based partitioning and packing optimization of a printable model}.
\newblock \bibinfo{journal}{\emph{{ACM} Transactions on Graphics}} \bibinfo{volume}{34}, \bibinfo{number}{6} (\bibinfo{date}{nov} \bibinfo{year}{2015}), \bibinfo{pages}{1--11}.
\newblock
\urldef\tempurl%
\url{https://doi.org/10.1145/2816795.2818064}
\showDOI{\tempurl}


\bibitem[Zhao et~al\mbox{.}(2022)]%
        {Zhao2020}
\bibfield{author}{\bibinfo{person}{Hang Zhao}, \bibinfo{person}{Qijin She}, \bibinfo{person}{Chenyang Zhu}, \bibinfo{person}{Yin Yang}, {and} \bibinfo{person}{Kai Xu}.} \bibinfo{year}{2022}\natexlab{}.
\newblock \bibinfo{title}{Online 3D Bin Packing with Constrained Deep Reinforcement Learning}.
\newblock
\newblock
\showeprint[arxiv]{2006.14978}~[cs.LG]


\bibitem[Zhao and Xu(2022)]%
        {zhao2022PCT}
\bibfield{author}{\bibinfo{person}{Hang Zhao} {and} \bibinfo{person}{Kai Xu}.} \bibinfo{year}{2022}\natexlab{}.
\newblock \showarticletitle{Learning Efficient Online 3D Bin Packing on Packing Configuration Trees}. In \bibinfo{booktitle}{\emph{International Conference on Learning Representations}}.
\newblock
\urldef\tempurl%
\url{https://openreview.net/forum?id=bfuGjlCwAq}
\showURL{%
\tempurl}


\bibitem[Zhao et~al\mbox{.}(2021)]%
        {Zhao2021}
\bibfield{author}{\bibinfo{person}{Hang Zhao}, \bibinfo{person}{Chenyang Zhu}, \bibinfo{person}{Xin Xu}, \bibinfo{person}{Hui Huang}, {and} \bibinfo{person}{Kai Xu}.} \bibinfo{year}{2021}\natexlab{}.
\newblock \showarticletitle{Learning practically feasible policies for online 3D bin packing}.
\newblock \bibinfo{journal}{\emph{Science China Information Sciences}} \bibinfo{volume}{65}, \bibinfo{number}{1} (\bibinfo{date}{dec} \bibinfo{year}{2021}).
\newblock
\urldef\tempurl%
\url{https://doi.org/10.1007/s11432-021-3348-6}
\showDOI{\tempurl}


\bibitem[Zou et~al\mbox{.}(2016)]%
        {Zou2016}
\bibfield{author}{\bibinfo{person}{Changqing Zou}, \bibinfo{person}{Junjie Cao}, \bibinfo{person}{Warunika Ranaweera}, \bibinfo{person}{Ibraheem Alhashim}, \bibinfo{person}{Ping Tan}, \bibinfo{person}{Alla Sheffer}, {and} \bibinfo{person}{Hao Zhang}.} \bibinfo{year}{2016}\natexlab{}.
\newblock \showarticletitle{Legible compact calligrams}.
\newblock \bibinfo{journal}{\emph{{ACM} Transactions on Graphics}} \bibinfo{volume}{35}, \bibinfo{number}{4} (\bibinfo{date}{jul} \bibinfo{year}{2016}), \bibinfo{pages}{1--12}.
\newblock
\urldef\tempurl%
\url{https://doi.org/10.1145/2897824.2925887}
\showDOI{\tempurl}


\end{thebibliography}

\appendix

\clearpage

\begin{figure*}[tb]
\centering
\includegraphics[width=0.85\linewidth]{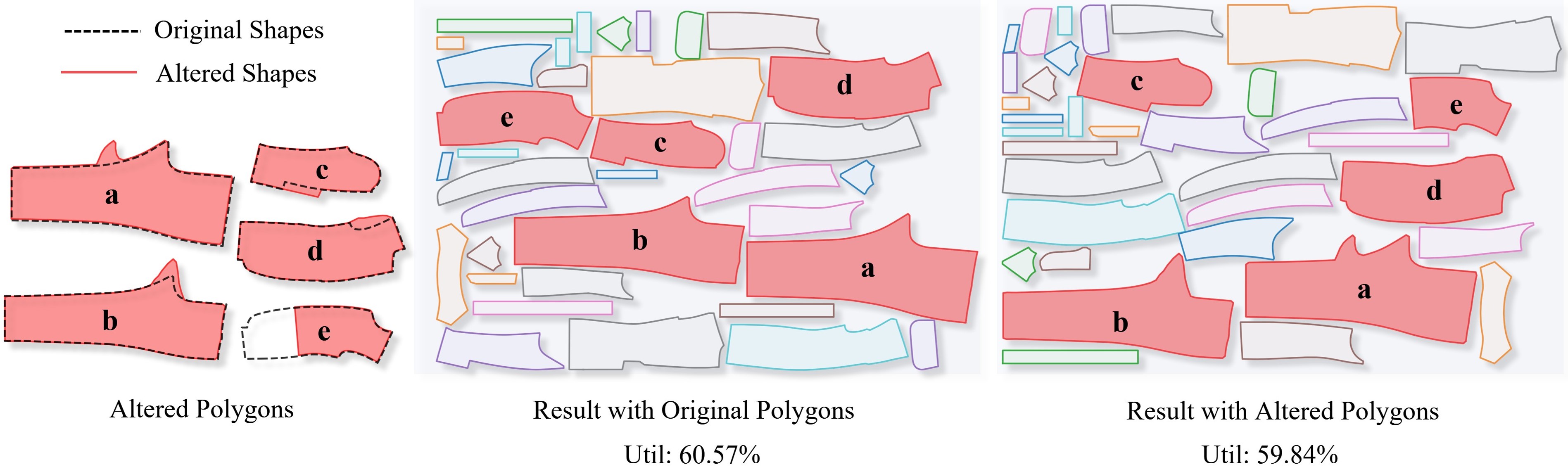}
\caption{The manually altered polygons (left), the results generated from the original polygons (middle), the results generated from the altered polygons (right). These polygons were respectively subjected to (a) the addition of similar geometric structures, (b) local geometric structure enlargement, (c-d) filling, and (e) cropping.}
\label{fig:shape}
\end{figure*} 

\begin{figure*}[tb]
\centering
\includegraphics[width=\linewidth]{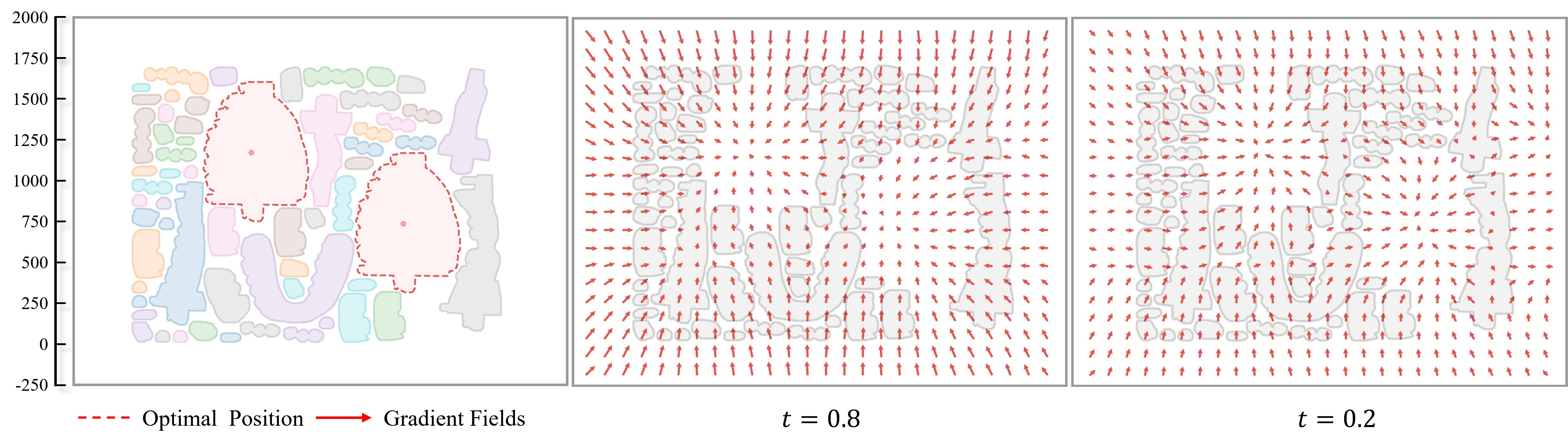}
\caption{ Visualization of gradient fields. }
\label{fig:gradient_field}
\end{figure*}

\begin{figure*}[tb]
\centering
\includegraphics[width=0.85\linewidth]{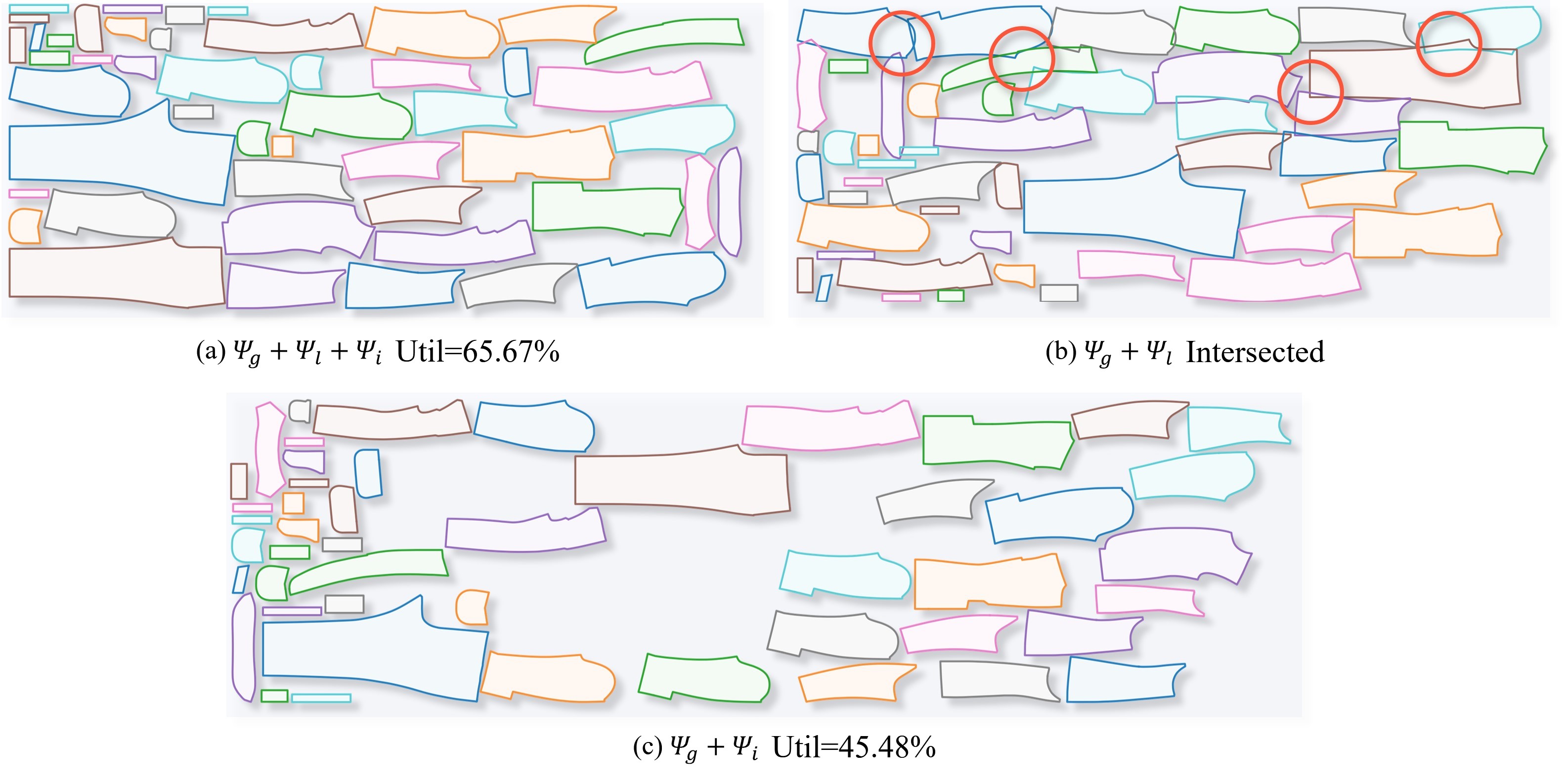}
\caption{ Ablation results. (a) All three network layers are activated, yielding satisfactory results. (b) Only the \emph{global layer} and \emph{local layer} are activated, resulting in overlapping outcomes. (c) Only the \emph{global layer} and \emph{intersection layer} are activated, producing non-overlapping outcomes with low space utilization. }
\label{fig:ablation}
\end{figure*} 

\begin{figure*}[tb]
\centering
\includegraphics[width=\linewidth]{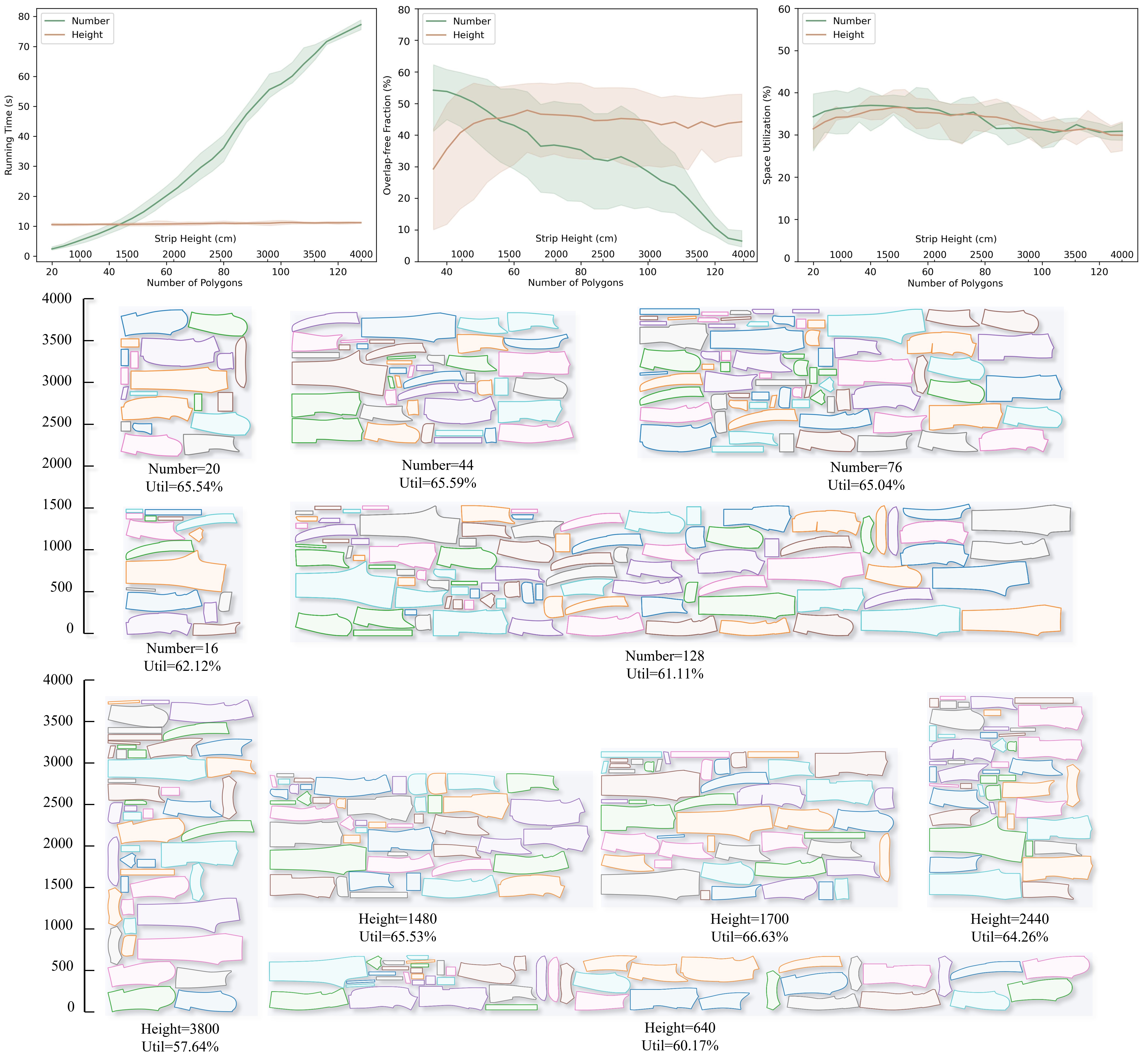}
\caption{ The figure illustrates the time consumption, overlap-free fraction, and space utilization achieved by our method ($b$=256, $E_s$=64) for varying numbers of polygons (20 to 128) and different strip heights (640 to 3920).  The shaded area indicates the range of maximum and minimum values of multiple test sets under the same x-axis value. Additionally, it visualizes a selection of example results.}
\label{fig:scalability}
\end{figure*}

\end{document}